\documentclass[lettersize,journal]{IEEEtran}
\usepackage{amsmath,amsfonts}
\usepackage{array}
\usepackage{subfig}
\usepackage{textcomp}
\usepackage{stfloats}
\usepackage{url}
\usepackage{verbatim}
\usepackage{graphicx}
\usepackage{cite}
\usepackage{bbding}
\usepackage{booktabs}
\usepackage{color}
\usepackage{xcolor}
\usepackage{colortbl}
\usepackage{amsfonts}
\usepackage{amsmath}
\usepackage[ruled,linesnumbered]{algorithm2e}
\usepackage{newfloat}
\usepackage{listings}

\begin{document}

\title{Learning Visual Conditioning Tokens to Correct Domain Shift for Fully Test-time Adaptation}

\author{Yushun Tang, Shuoshuo Chen, Zhehan Kan, Yi Zhang, Qinghai Guo, and Zhihai He,~\IEEEmembership{Fellow, IEEE}
        % <-this % stops a space
\thanks{
Y. Tang, S. Chen, Z. Kan, Y. Zhang, and Z. He are with the Department of Electrical and Electronic Engineering, Southern University of Science and Technology, Shenzhen, China. Z. He is also with the Pengcheng Lab, Shenzhen, China. Q. Guo is with the Advanced Computing and Storage Laboratory, Huawei Technologies Co., LTD.}

\thanks{
Corresponding author: Z. He 
(hezh@sustech.edu.cn).}% 

}

\markboth{IEEE Transactions on Multimedia}%
{Shell \MakeLowercase{\textit{et al.}}: A Sample Article Using IEEEtran.cls for IEEE Journals}


\maketitle

\begin{abstract}
Fully test-time adaptation aims to adapt the network model based on sequential analysis of input samples during the inference stage to address the cross-domain performance degradation problem of deep neural networks. This work is based on the following interesting finding: in transformer-based image classification, the class token at the first transformer encoder layer can be learned to capture the domain-specific characteristics of target samples during test-time adaptation. This learned token, when combined with input image patch embeddings, is able to gradually remove the domain-specific information from the feature representations of input samples during the transformer encoding process, thereby significantly improving the test-time adaptation performance of the source model across different domains. We refer to this class token as \textit{visual conditioning token} (VCT). To successfully learn the VCT, we propose a bi-level learning approach to capture the long-term variations of domain-specific characteristics while accommodating local variations of instance-specific characteristics. Experimental results on the benchmark datasets demonstrate that our proposed bi-level visual conditioning token learning method is able to achieve significantly improved test-time adaptation performance by up to 1.9\%. 
% The proposed method can be applied to other transformer-based models for computer vision tasks beyond image classification. 
% Our code is available at {https://anonymous.4open.science/r/VCT-C9D4}.
\end{abstract}


\section{Introduction}
\label{sec: Introduction}
Although deep convolutional neural networks (CNNs) and transformers have achieved remarkable success in various machine learning tasks, their performance tends to deteriorate significantly in the presence of data distribution shifts~\cite{quinonero2008dataset} between the training data in the source domain and the testing data in the target domain \cite{ mirza2021robustness, wang2022cross,active_zhou,guided_zhang,meng2022dual,deng2023cross,chen2024learning,zhang2024hiker}. The deployment of deep models in test domains with distribution shifts remains a challenging task but with important applications in practice. 
Source-free unsupervised domain adaptation (UDA) methods~\cite{liang2020we,Wang_2022_CVPR,li2020model,limttn,tang2023cross, tian2022vdmda} aim to adapt network models without access to the source-domain samples. However, these methods need to access to the entire target dataset, which is impractical in real-world application scenarios.

Recently developed test-time adaptation (TTA) methods have the potential to adapt a pre-trained model to unlabeled data during testing, before making predictions \cite{Sunttt,wang2020tent,mirza2022norm,niu2022efficient,wang2022continual,tang2023neuro,yuan2023robust,wen2023test,kan2023self}. There are two major categories of approaches for TTA: (1) test-time training (TTT) \cite{Sunttt,liu2021ttt++,gandelsman2022test} and (2) fully test-time adaptation \cite{wang2020tent,mirza2022norm,niutowards}, depending on whether the source domain data is accessed or not. The TTT methods methods often require proxy tasks or special training in the source domain. However, in practical applications, access to the source data may not be available. In this work, we focus on the fully test-time adaptation, which does not rely on source domain data.

Existing methods for fully test-time adaptation mainly focus on utilizing the entropy loss to adapt the model based on back-propagation. 
For example, the TENT method \cite{wang2020tent} updates the batch normalization module by minimizing an entropy loss. The DUA \cite{mirza2022norm} method adapted the moving average decay factor of the normalization layer by using only a small amount of target domain data and its augmentations.
The MEMO method \cite{zhang2022memo} optimizes the entropy of the averaged prediction over multiple random augmentations of the input sample.
The VMP method \cite{jing2022variational} introduces perturbations into the model parameters based on variational Bayesian inference.
Note that these methods assume that the label distribution of each batch in the target domain is well-balanced, which is however not the case in practice.
To address this issue, the SAR method \cite{niutowards} proposes a sharpness-aware and reliable optimization scheme, which removes samples with large gradients and encourages model weights to converge to a flat minimum.
The TTN method \cite{limttn} optimizes the interpolating weight during the post-training phase but has access to the labeled source data. The RoTTA method \cite{yuan2023robust} proposes a robust batch normalization by category-balanced sampling with timeliness and uncertainty.

\begin{figure*}[!htbp]
\setlength{\abovecaptionskip}{0.0cm}
\setlength{\belowcaptionskip}{-0.5cm}
    \centering
    \includegraphics[width = 0.93\textwidth]{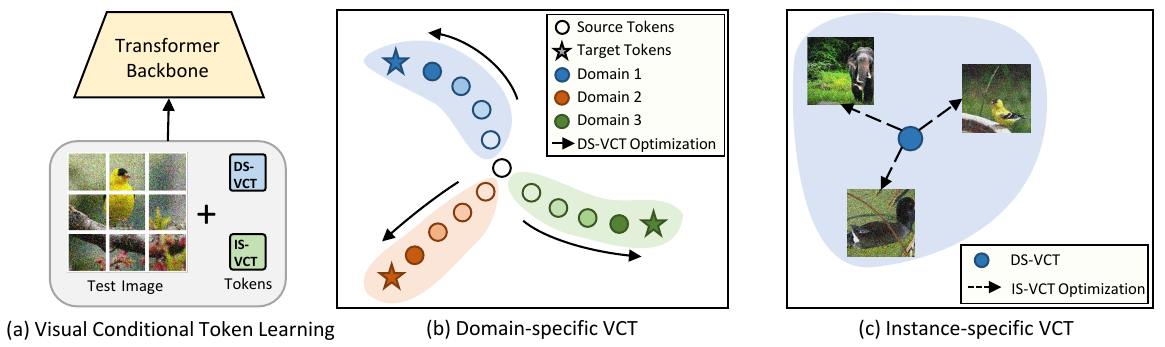} 
    \caption{An illustration of our proposed bi-level visual conditioning token learning. (b): The adaptation process of the domain-specific visual conditioning token (DS-VCT), which aims to learn domain-specific information for all samples in the same domain.  (c): The adaptation process of the instance-specific visual conditioning token (IS-VCT), which aims to learn instance-specific information for each target sample based on DS-VCT.}
    \label{fig: intro}
% \vspace{-0.5cm}
\end{figure*}

These methods primarily utilize architectures from the ResNet family \cite{he2016deep,zagoruyko2016wide}. Recently, transformer-based methods \cite{vaswani2017attention,dosovitskiyimage,tang2022gloss,he2022revisiting} have achieved remarkable success in various machine-learning tasks. However, the TTA for transformer architecture has been seldom studied. In this work, we use the Vision Transformer (ViT) as the backbone encoder for our image classification task. Learning task-specific prompts to improve the performance in downstream tasks has been studied in recent literature \cite{jia2022visual,gan2023decorate,DBLP:conf/emnlp/LesterAC21, liu2021ptuningv2, li2021prefix}. In target datasets like ImageNet-C \cite{hendrycks2018benchmarking}, domain-specific corruptions can lead to significantly performance degradation by perturbing the semantic features of the samples.
We recognize that the key task in test-time adaptation is to learn the domain-specific information and separate the domain-specific information from input sample features. 
For the ViT encoder, the class token pre-trained on the source domain captures the semantic information of source images. Due to the distribution shift, this pre-trained token may not be able to generalize to the target images, since the prior knowledge on semantics embedding in the class token is source-specific and the pre-trained class token has no prior knowledge about the domain shift that it encounters during test time.

In this work, we find that the class token at the first transformer encoder layer can be learned to capture the domain shift characteristics experimentally during test-time adaptation. We refer to this token as \textit{visual conditioning token} (VCT).
Once successfully learned, the VCT is able to perform conditioning operations on the input image to correct image perturbations caused by domain shift. With the domain shift perturbations or corruptions being removed from the image features, the network model will be more robust to domain changes, resulting in significantly improved generalization performance. 
We observe that, at the domain level, different distribution shifts have different impacts on the overall model performance of the test dataset. Furthermore, at each individual image level, the performance impacts on different images by the distribution shift are different.  
Motivated by this, to successfully learn the VCT, we propose a bi-level learning approach to 
characterize long-term variations of domain-specific characteristics and local variations of instance-specific characteristics of the 
domain shift.
As illustrated in Fig. \ref{fig: intro}, the VCT is split into two parts, the domain-specific VCT (DS-VCT) and instance-specific VCT (IS-VCT), 
aiming to characterize the impacts of domain shift at the whole test dataset (or domain level) and the individual image level. 
The long-term domain-specific VCT is updated at each batch and keeps continual updating during inference in the target domain, while the short-term instance-specific VCT is updated at each sample and reset to zero at the end of each batch.  It learns the instance-specific knowledge for current samples and forgets the prior bias dynamically. Finally, these two learnable tokens are embedded into the source pre-trained ViT for fully test-time adaptation. The design  proposed in this work aligns well with the principles of complementary learning theory \cite{mcclelland1995there,kumaran2016learning} in neuroscience.
Experimental results on multiple datasets demonstrate that our proposed method can significantly improve the adaptation performance of transformer-based network models and outperform existing state-of-the-art methods by large margins.

\vspace{-0.2cm}
\section{Related Work and Major Contributions}
\label{sec: Related}
This work is related to existing work on test-time adaptation, prompt learning, and complementary learning. 

\paragraph{Test-time Adaptation}
Test-time adaptation aims to online adapt the trained model while testing the input samples in the target domain. \cite{DBLP:conf/icml/SunWLMEH20} proposed Test-time Training (TTT) by optimizing a self-supervised loss through a proxy task on the source before adapting to the target domain.
TTT++~\cite{liu2021ttt++} encouraged the test feature distribution to be close to the training one by matching the moments estimated online. It should be noted that this method requires specific training on the source data, which is not available in fully test-time adaptation scenarios \cite{wang2020tent}.
TENT~\cite{wang2020tent} fine-tuned the scale and bias parameters of batch normalization layers using an entropy minimization loss during the inference process.
% DUA
DUA~\cite{mirza2022norm} adapted the statistics of batch normalization layer only on a tiny fraction of test data and augmented a small batch of target data to adapt the model.
% improvmen (ECCV2022) 
\cite{choi2022improving} proposed a shift-agnostic weight regularization and an auxiliary task for the alignment between the source and target features.  Note that this method requires the source data for computing the source prototypes.
% continual cotta (cvpr2022)+ gradual continual(nips2022)
The continual test-time adaptation methods \cite{wang2022continual, marsden2022gradual} consider online TTA where target data is continually changing during inference. 
% parameter-free (cvpr2022)
Instead of using parameters of the pre-trained model, \cite{boudiaf2022parameter} only adapted the model's output by optimizing an objective function based on  Laplacian adjusted maximum-likelihood estimation. 
% Besides image classification, test-time adaptation has been successfully applied in various machine learning tasks, such as scene deblurring \cite{chi2021test}, super-resolution \cite{shocher2018zero}, human pose estimation \cite{li2021test}, image segmentation~\cite{hu2021fully}, object detection \cite{mirza2022norm}, etc. 

\paragraph{Prompt Learning}
The concept of prompt originated from the field of Natural Language Processing (NLP), where the linguistic instructions, called prompts, are appended to the input text to guide the pre-trained language model to perform specific downstream tasks \cite{liu2023promptsurvey}. Recent approaches \cite{DBLP:conf/emnlp/LesterAC21, liu2021ptuningv2, li2021prefix} have suggested representing prompts as task-specific vector inputs and optimizing them directly using error back-propagation. Such prompt tuning methods learn the prompts from downstream data in the input embedding space, requiring much fewer parameters to be updated during the adaptation process.

Recently, in the realm of Vision-Language Models, several approaches have emerged that leverage prompt tuning to enhance performance \cite{zhou2022coop, zhou2022conditional, ma2023prompt, shu2022test, ge2022DAPL,zhang2024concept,zhang2024cross,zhang2023bdc}. CoOp \cite{zhou2022coop} introduced a method that tunes the prompt for the text encoder in CLIP \cite{radford2021clip}, setting a precedent for prompt-based optimization. Extending this concept, CoCoOp \cite{zhou2022conditional} further conditioned the prompt learning on the input data, aiming to mitigate the out-of-distribution challenge. 
While these innovative methods primarily target the text encoder in Vision-Language Models, our approach stands apart by solely focusing on enhancing the vision transformer (ViT) encoder, exploring a distinct path for prompt-based learning in the visual domain..

% ======Vision prompt=======
Beyond text prompt learning, recent advancements have explored visual prompts for enhancing computer vision tasks. For instance, the method by \cite{gan2023decorate} learns both domain-specific and domain-agnostic prompts that are pixel-wise attached to input images for continuous Test-Time Adaptation (TTA) tasks. BlackVIP \cite{oh2023blackvip} takes a different approach, learning individualized prompts for each image that share the same shape as the input, without requiring knowledge of the pre-trained model's architecture or parameters.
Visual Prompt Tuning (VPT) \cite{jia2022visual} introduced a technique that prepends task-specific learnable parameters to the input sequence of each Vision Transformer (ViT) encoder layer, while keeping the pre-trained encoder backbone frozen. This approach has been successfully applied in image synthesis tasks \cite{sohn2022visualgan} and extended to test-time adaptation with DePT \cite{gao2022visual}, leveraging a memory bank for online pseudo-labeling. However, one notable limitation of VPT is the introduction of hundreds of new tokens, which significantly escalates the computational complexity of the self-attention mechanism due to its quadratic dependency on the number of input tokens.
All of these prompt tuning methods introduced additional learnable parameters.
Different from VPT with numerous extra tokens for prompt learning, we choose to adapt the existing one class token in the transformer encoder pre-trained on the source for test-time adaptation. 

\paragraph{Complementary Learning}
The proposed VCT method in this work aligns well with the principles in complementary learning theories \cite{mcclelland1995there,kumaran2016learning}
with long-term and short-term learning.
Though fully TTA requires the adaptation to only rely on the current batch of test samples, such consecutive update on a narrow portion of the distribution in the target domain can bring the risk of dramatically degrading the performance due to overspecialization~\cite{boudiaf2022parameter}. Biologically, the brain avoids such risk using the following mechanism: generalizing itself across perceived experiences while holding episodic memories for separated events~\cite{parisi2019continual}. The complementary learning systems (CLS) theory \cite{mcclelland1995there,kumaran2016learning} is based on such a learning mechanism. 
In CLS, the neocortex performs slow learning of structured knowledge and the hippocampus performs fast learning of episodic information. More specifically, the hippocampus uses a large learning rate to allow an event being encoded in episodic memory. Conversely, the neocortex uses a small learning rate to extract generalities and build overlapping representations of the learned knowledge~\cite{o2004division}. Therefore, the collaboration between the neocortical and hippocampal brain components is crucial for concurrently learning regularities (statistics of the environment) and specifics (episodic memories)~\cite{gais2007sleep,kitamura2017engrams}. It has been demonstrated to be effective in mitigating catastrophic forgetting in continually learning \cite{aranilearning, yu2023continual}. Biological evidence shows that such short-term and long-term memories are achieved through synaptic plasticity~\cite{neves2008synaptic}. 
These brain-inspired complementary learning methods utilize the collaboration of long-term and short-term learning to develop more effective representations.
Motivated by this slow and fast learning mechanism, we propose long-term domain-specific and short-term instance-specific visual conditioning token learning for effective test time adaptation.

\paragraph{Major Contributions}
Compared to existing work, the major contributions of this work can be summarized as: (1) We have made an interesting finding that the class token at the first layer of the transformer encoder can be used as a target for optimization during test-time adaptation. It is able to capture the domain-specific characteristics of test samples in the target domain and is able to gradually correct the domain shift perturbations in the test samples.
(2) We develop a bi-level approach to effectively learn the visual conditioning token which is able to capture the long-term variations across domains while being able to accommodate the local variations across individual samples. 
(3) Experimental results demonstrate that our proposed bi-level visual conditioning token learning method is able to significantly improve TTA performance by up to 1.9\%.

\section{Method}
\label{sec: Method}
In this section, we present our method of bi-level visual conditioning token learning for fully test-time adaptation.

\vspace{-0.3cm}
\subsection{Method Overview}
Suppose that a model $\mathcal{M}=f_{\theta_s}(y|X_s)$ with parameters $\theta_s$ has been successfully trained on the source data $\{X_s\}$ with labels $\{Y_s\}$ where the true distribution is $p_s(y|X_s)$. 
During fully test-time adaptation, we are given the target data $\{X_t\}$ with unknown labels $\{Y_t\}$. Our goal is to adapt the trained model in an unsupervised manner during testing. Given a sequence of input sample batches $\{\mathbf{B}_1, \mathbf{B}_2, ..., \mathbf{B}_J\}$ from $\{X_t\}$, the $j$-th adaptation of the network model can only rely on the $j$-th batch of test samples $\mathbf{B}_j$. We follow the \textit{wild} test-time adaptation setting in SAR \cite{niutowards} where the mini-batch $\mathbf{B}_j$ can contain only one sample or samples in the mini-batch can be imbalanced.

\begin{figure*}[!ht]
\setlength{\abovecaptionskip}{0.0cm}
\setlength{\belowcaptionskip}{-0.4cm}
    \centering
    \includegraphics[width = 0.96\textwidth]{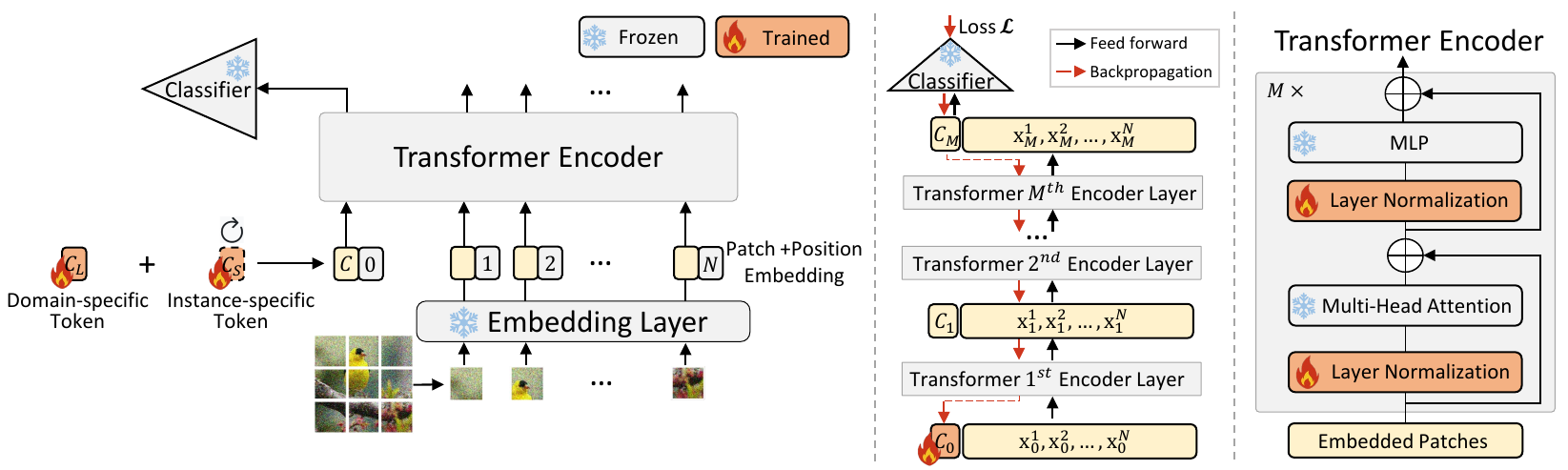}
    \caption{An overview of the proposed VCT method. During inference in the target domain, the class token which consists of the domain-specific visual conditioning token (DS-VCT) and the instance-specific visual conditioning token (IS-VCT), is updated before making a prediction given each mini-batch testing sample. The IS-VCT and its gradient are reset after each prediction (\textbf{Left}). The gradient flow in the multi-layer transformer encoder (\textbf{Center}). The details of each encoder layer (\textbf{Right}).}
    \label{fig: framework}
% \vspace{-0.5cm}
\end{figure*}

In this work, we learn the class token at the first transformer encoder layer to capture the domain characteristics during test-time adaptation. This class token, 
called the visual conditioning token in this work, is combined with the input patch embeddings and fed into the transformer encoder. It is able to gradually correct the domain-specific information from the input image feature during the transformer encoding and inference process, thereby significantly improving the test-time adaptation performance of the source model across different domains.  In the following sections, we explain the proposed VCT approach and its learning process in more details. 

\subsection{Learning the Visual Conditioning Token}
Following the SAR method in \cite{niutowards}, we use the ViT as our backbone network. 
In ViT, the input image is divided into $N$ patches
$\mathbf{P} = \{\mathbf{P}^1, \mathbf{P}^2, \cdots, \mathbf{P}^N \}$.
Each patch $\mathbf{P}^i$ is encoded into a patch embedding $\mathbf{x}^i$.  
A class token $\mathbf{C}$, which is a learnable parameter vector of size $d$, is concatenated with all patch embeddings of the image and then fed into an $M$-layer 
transformer encoder $\mathcal{T}$. At each layer $\mathcal{T}_m$, $1\le m\le M$, the class token and the patch embeddings are updated based on the multi-head self-attention mechanism and denoted as 
$[\mathbf{C}_m;\   \mathbf{x}^1_m, \mathbf{x}^2_m, \cdots, \mathbf{x}^N_m]$. We have 
\begin{equation}
\begin{split}
 &[\mathbf{C}_m;\   \mathbf{x}^1_m, \mathbf{x}^2_m, \cdots, \mathbf{x}^N_m] \\
 = \mathcal{T}_m(&[\mathbf{C}_{m-1};\   \mathbf{x}^1_{m-1}, \mathbf{x}^2_{m-1}, \cdots, \mathbf{x}^N_{m-1}]).   
\end{split}
\end{equation}
At the last layer, the output class token $\mathbf{C}_M$ is fed into an MLP network to predict the class label for the input image \cite{dosovitskiyimage}.
It has been observed that the class token $\mathbf{C}_M$ acts as a global description vector that captures the overall semantics of the image and helps the model to make accurate predictions \cite{zhang2021token,xu2022multi}.
During model training in the source domain, the class token 
$\mathbf{C}_0$ at the first layer is initialized as a random vector. 
During the training process, $\mathbf{C}_0$ is updated based on the gradients of the error propagation from the upper layers of the encoder as shown in the center of Fig. \ref{fig: framework}.

In this work, interestingly, we find that the class token $\mathbf{C}_0$ at the input layer of the transformer encoder, if effectively learned, plays an important role in characterizing the domain shift perturbations in the test samples. 
During training in the source domain, the ViT-based classifier needs to learn the following probability function of output $y$ for the input image $\mathbf{X}$
\begin{equation}
    p(y) = P(y \ |\  \mathbf{C}_0, \mathbf{X}).
\end{equation}
Here, the VCT $\mathbf{C}_0$ remains invariant across all test classes and samples.
During training in the source domain, 
the class token $\mathbf{C}_0$ is learned to minimize the cross-entropy loss of all training samples
\begin{equation}
    \mathbf{C}_0^* = \arg \min \limits_{\mathbf{C}_0} \sum_{\mathbf{X}\in \{X_s\}} 
    \mathcal{L}_{CE}[P(y \ |\  \mathbf{C}_0, \mathbf{X}), y_s],
\end{equation}
where $y_s$ is the ground-truth label of the input image $\mathbf{X}$, and $\mathcal{L}_{CE}[\cdot,\cdot]$ represents the cross-entropy loss.
From this perspective, we can see that $\mathbf{C}_0$ captures some characteristics of the whole set of training samples in the source domain. In other words, it contains domain-specific information about the source samples.
It can be considered as a prior distribution over all class labels, which encodes  prior knowledge about the whole set of training data. 
Note that, in existing transformer-based methods,  once trained in the source domain, the class token $\mathbf{C}_0$ is frozen during inference in the target domain. 
Also, it was rarely used in existing machine learning methods, except being used as a placeholder to learn the final class token $\mathbf{C}_M$ to capture the global semantic information for image classification. 

In this work, we find that $\mathbf{C}_0$ can be used as the target for optimization during test-time adaptation to re-capture the global characteristics of samples in the target domain.
Once combined with image patch embeddings, it is able to correct the impact of domain shift on the classification performance in the target domain to achieve improved generalization capability of the source model. Therefore, we refer to  $\mathbf{C}_0$ 
as the \textbf{visual conditioning token}, which performs some cleanup treatment of the test samples. 
During the test-time adaptation process, 
we update the parameter weights of the visual conditioning token $\theta_{C0}$ using back-propagation with  loss $\mathcal{L}(\theta_t; x)$ introduced in Section \ref{sec: multi} as:
\begin{equation}\label{eq: LTPP_update}
    \tilde{\theta}_{C0} = \theta_{C0}-\eta \cdot \nabla_{\theta_{C0}} \mathcal{L}(\theta_t; x),
\end{equation}
where $\eta$ is learning rate and $\nabla_{\theta_{C0}}$ are gradients of the token.

\subsection{Bi-Level Visual Conditioning Token}
\label{sec: multi}
In the above section, we have learned the VCT to capture the domain-specific characteristics which should remain relatively invariant across samples from different classes. 
From the test-time adaptation perspective, this domain-specific VCT 
captures the overall characteristics of the domain shift and its long-term average impact on the test dataset. 
We observe that the impact of domain shift also occurs at each individual image level. 
From one batch of samples to the next, the sample characteristics often vary significantly since they are from different classes. This local sample variation will result in local variations in the error gradients and therefore in the update of the VCT vector. 
To further improve the performance of test-time adaptation, it is necessary to introduce a short-term VCT to capture instance-specific characteristics and accommodate this local variation.
Specifically, the proposed VCT for the $n$-th sample in a mini-batch can be written as 
\begin{equation}
    \mathbf{C}_0[n] = \mathbf{C}_L + \mathbf{C}_S[n],  \quad 0 \le n < B,
\end{equation}
where $B$ is the batch size. Here, $\mathbf{C}_L$ is the long-term domain-specific token shared by all samples in a mini-batch that captures the global context of the test domain.
$\mathbf{C}_S$ is the short-term instance-specific token that captures the local variations of individual image instances.

To effectively learn these two VCTs, we propose a bi-level learning approach. Specifically, the long-term domain-specific token $\mathbf{C}_L$ is initialized with the class token vector from the pre-trained source model and updated slowly at each batch throughout the test-time adaptation process. 
On the other hand, the short-term instance-specific token $\mathbf{C}_S$  is initialized with zero at the start of each batch and updated before prediction. This allows the short-term instance-specific token to quickly adapt to the local details of the input image.
More importantly, it is reset at the end of each batch to safeguard against overfitting of the current batch instances, which could potentially impact the performance of subsequent batches.

Given the mini-batch $\mathbf{B}_j$, following the aforementioned idea,  our proposed method aims to minimize the loss function $\mathcal{L}(\theta_t; x)$ with respect to the learnable weights $\theta_t$, which include tokens $\mathbf{C}_L$  and $\mathbf{C}_S$, as well as the affine parameters of layer normalization modules. The weights of long-term domain-specific token $\theta_L$ and short-term instance-specific token $\theta_S$ are updated as follows:
\begin{equation}\label{eq: L_VCT_update}
    \tilde{\theta_L} = {\theta_L}-\eta_l \cdot \nabla_{\theta_L} \mathcal{L}(\theta_t; x),
\end{equation}
\begin{equation}\label{eq: S_VCT_update}
    \tilde{\theta_S} = {\theta_S}-\eta_s \cdot \nabla_{\theta_S} \mathcal{L}(\theta_t; x),
\end{equation}
where $\eta_l$, and $\eta_s$ are the learning rates of the domain-specific token $\mathbf{C}_L$ and instance-specific token $\mathbf{C}_S$, respectively, and $\nabla_\theta$ are the corresponding gradients.

The loss function $\mathcal{L}(\theta_t; x)$ in test-time adaptation is often defined by the entropy of the current batch. In practice, we find such minimization will cause model collapse. To address this issue, we use the reliable entropy minimization and sharpness-aware entropy minimization methods in \cite{niutowards}.
The reliable entropy minimization filters out testing samples with relatively large entropy to reduce the impact of noisy samples on the model's fine-tuning and makes it more robust to incomplete or noisy data. The sharpness-aware entropy minimization encourages the model weights to converge to a flat minimum, indicating that the model is robust to small perturbations in the weights. The overall optimization loss is defined as:
\begin{equation}\label{eq: loss}
     \mathcal{L}(\theta_t; x) = \mathbb{I}[\mathbb{E}(\theta_t; x)<E_0]\cdot \mathbb{E}(\theta_t; \mathbf{B}_j),
\end{equation}
where $\mathbb{I}[\mathbb{E}(\theta_t; x)<E_0]$ is the mask to filter out test samples when entropy is larger than the threshold $E_0$, $\mathbb{E}$ is the entropy function. The pseudo-code of our proposed VCT method is shown in Algorithm \ref{alg: algorithm}.

\begin{algorithm}[!htbp]
\caption{Pseudo code of the proposed algorithm.}
\label{alg: algorithm}
\KwIn {Source pre-trained model $f_{\theta_s}$; target dataset $\{X_t\}$.}
\KwOut {The prediction of target samples $\{\hat{\mathbf{y}}\}$.}
Initialize the testing model $f_{\theta_t}$ with the source pre-trained model $f_{\theta_s}$ parameter weights; initialize the bi-level VCTs $\theta_{L}/\theta_{S}$ with the source class token and zero respectively; add all parameters $\tilde{\theta}$ to be fine-tuned to the Sharpness Aware Minimization optimizer; learning rate $\eta_s$, $\eta_l > 0$; \\
\For {batch $\mathbf{B}_j$ \textbf{in} $\{X_t\}$}{
 Compute the gradient $\nabla g$ with the loss function in Eqn. (8)\;
 Update $\tilde{\theta_L} \leftarrow \tilde{\theta_L}-\eta_l \nabla g_l$; $\tilde{\theta_S} \leftarrow \tilde{\theta_S}-\eta_s \nabla g_s$\;
 Output $\hat{\mathbf{y}}$ =  $f_{\theta_t}(\mathbf{B}_j)$\;
 Reset $\tilde{\theta_S} = \mathbf{0}$, $\nabla g_s = \mathbf{0}$\ \tcp*{Reset the IS-VCT and its gradient}}
\end{algorithm}

\begin{table*}[!htbp]
\setlength{\abovecaptionskip}{0.1cm}
\setlength{\belowcaptionskip}{-0.2cm}
\begin{center}
\caption{Top-1 Classification Accuracy (\%) for each corruption in \textbf{ImageNet-C} under \textbf{Normal} setting. The best result is shown in \textbf{bold}.} 
\label{table: normal}
\resizebox{\linewidth}{!}
{
\begin{tabular}{l|ccccccccccccccc|c}
\toprule
Method & gaus & shot & impul & defcs & gls & mtn & zm & snw & frst & fg & brt & cnt & els & px & jpg & Avg.\\	
\midrule
\multicolumn{17}{c}{Level 3} \\
\midrule
Source & 51.6 & 46.9 & 50.5 & 48.7 & 37.2 & 54.7 & 41.6 & 35.1 & 33.5 & 67.8 & 69.3 & 74.8 & 65.8 & 66.0 & 63.7 & 53.8 \\
MEMO~\cite{zhang2022memo} & 61.9 & 57.7 & 61.4 & 57.0 & 45.4 & 61.8 & 49.8 & 46.6 & 43.1 & 73.9 & 75.7 & 79.6 & 72.6 & 72.1 & 70.5 & 61.9\\
DDA~\cite{gao2022back} & 59.8 & 58.2 & 59.5 & 43.4 & 43.2 & 50.4 & 40.9 & 34.2 & 34.3 & 55.2 & 64.9 & 64.0 & 64.2 & 63.7 & 62.8 & 53.2\\
TENT~\cite{wang2020tent} & 67.0 & 66.1 & 66.2 & 66.4 & 60.8 & 69.0 & 61.1 & 65.0 & 60.4 & 75.1 & 78.1 & 78.8 & 74.7 & 75.8 & 72.4 & 69.1 \\
SAR~\cite{niutowards} & 66.7 & 66.0 & 66.1 & 66.4 & 60.9 & 68.7 & 61.3 & 64.9 & 63.4 & 75.0 & 77.9 & 78.7 & 74.5 & 75.6 & 72.3 & 69.2 \\ 
\rowcolor{gray!20}
\textbf{Ours} & \textbf{67.2} & \textbf{66.4} & \textbf{66.6} & \textbf{66.8} & \textbf{61.8} & \textbf{69.3} & \textbf{61.9} & \textbf{65.5} & \textbf{63.9} & \textbf{75.5} & \textbf{78.2} & \textbf{79.1} & \textbf{75.0} & \textbf{75.9} & \textbf{72.9} & \textbf{69.7} \\ 
\rowcolor{gray!20}
&$_{\pm0.1}$ &$_{\pm0.1}$ &$_{\pm0.0}$&$_{\pm0.0}$&$_{\pm0.1}$&$_{\pm0.0}$&$_{\pm0.1}$&$_{\pm0.2}$&$_{\pm0.1}$&$_{\pm0.0}$&$_{\pm0.0}$&$_{\pm0.0}$&$_{\pm0.1}$&$_{\pm0.0}$&$_{\pm0.1}$&$_{\pm0.0}$ \\
\midrule
\multicolumn{17}{c}{Level 5} \\
\midrule
Source & 9.5 & 6.8 & 8.2 & 29.0 & 23.5 & 33.9 & 27.1 & 15.9 & 26.5 & 47.2 & 54.7 & 44.1 & 30.5 & 44.5 & 47.8 & 29.9 \\
MEMO~\cite{zhang2022memo} & 21.6  & 17.3  & 20.6  & 37.1  & 29.6  & 40.4  & 34.4  & 24.9  & 34.7  & 55.1  & 64.8  & 54.9  & 37.4  & 55.4  & 57.6  & 39.1\\
DDA~\cite{gao2022back} & 41.3 & \textbf{41.1} & 40.7 & 24.4 & 27.2 & 30.6 & 26.9 & 18.3 & 27.5 & 34.6 & 50.1 & 32.4 & 42.3 & 52.2 & 52.6 & 36.1\\
TENT~\cite{wang2020tent} & 42.4 & 1.2 & 43.2 & 52.2 & 47.7 & 55.3 & 50.1 & 19.8 & 21.6 & 66.2 & \textbf{74.7} & 64.7 & 52.1 & 66.7 & 64.1 & 48.1 \\
SAR~\cite{niutowards} & 44.4 & 31.1 & 45.3 & 52.9 & 49.9 & 55.8 & 51.2 & \textbf{57.9} & 51.6 & 66.5 & 74.5 & 64.4 	& 55.4 	& 66.6 & 64.0 & 55.4 \\ 
\rowcolor{gray!20}
\textbf{Ours} & \textbf{45.0} & 35.9 & \textbf{45.9} & \textbf{53.5} & \textbf{51.0} & \textbf{56.6} & \textbf{51.8} & 57.3 & \textbf{53.3} & \textbf{67.5} & \textbf{74.7} & \textbf{64.9} & \textbf{56.4} & \textbf{67.0} & \textbf{64.7} & \textbf{56.4} \\ 
\rowcolor{gray!20}
&$_{\pm0.0}$&$_{\pm6.5}$&$_{\pm0.4}$&$_{\pm0.1}$&$_{\pm0.1}$&$_{\pm0.1}$&$_{\pm0.1}$&$_{\pm0.4}$&$_{\pm2.5}$&$_{\pm0.1}$&$_{\pm0.0}$&$_{\pm0.1}$&$_{\pm0.1}$&$_{\pm0.2}$&$_{\pm0.1}$&$_{\pm0.4}$ \\
\bottomrule
\end{tabular}
}
% \vspace{-5pt}
\end{center}
\end{table*}

\begin{table*}[!htbp]
\setlength{\abovecaptionskip}{0.1cm}
\setlength{\belowcaptionskip}{-0.2cm}
\begin{center}
\caption{Top-1 Classification Accuracy (\%) for each corruption in \textbf{ImageNet-C} under \textbf{Imbalanced label shifts}.} % at the highest severity (Level 5).
\label{table: imbalanced}
\resizebox{\linewidth}{!}
{
\begin{tabular}{l|ccccccccccccccc|c}
\toprule
Method & gaus & shot & impul & defcs & gls & mtn & zm & snw & frst & fg & brt & cnt & els & px & jpg & Avg.\\	
\midrule
\multicolumn{17}{c}{Level 3} \\
\midrule
Source & 51.5 & 46.8 & 50.4 & 48.7 & 37.1 & 54.7 & 41.6 & 35.1 & 33.3 & 68.0 & 69.3 & 74.9 & 65.9 & 66.0 & 63.6 & 53.8 \\
MEMO~\cite{zhang2022memo} & 62.1 & 57.9 & 61.5 & 57.2 & 45.6 & 62.0 & 49.9 & 46.5 & 43.1 & 74.1 & 75.8 & 79.7 & 72.6 & 72.3 & 70.6 & 62.1 \\
DDA~\cite{gao2022back} & 59.7 & 58.2 & 59.4 & 43.5 & 43.3 & 50.5 & 41.0 & 34.3 & 34.4 & 55.4 & 65.0 & 64.2 & 64.1 & 63.8 & 62.9 & 53.3\\
TENT~\cite{wang2020tent} & 68.7 & 68.0 & 68.1 & 68.2 & 63.8 & 70.9 & 63.8 & 67.6 & 41.9 & 76.3 & 78.8 & 79.5 & 75.9 & 76.7 & 73.7 & 69.5 \\
SAR~\cite{niutowards} & 68.8 & 68.2 & 68.4 & 68.3 & 64.7 & 71.0 & 64.2 & 68.1 & 66.0 & 76.4 & 79.0 & 79.6 & 76.2 & 77.1 & 74.1 & 71.3\\ 
\rowcolor{gray!20}
\textbf{Ours} & \textbf{69.2} & \textbf{68.6} & \textbf{68.7} & \textbf{68.5} & \textbf{65.2} & \textbf{71.3} & \textbf{64.7} & \textbf{68.3} & \textbf{66.4} & \textbf{76.8} & \textbf{79.1} & \textbf{79.8} & \textbf{76.6} & \textbf{77.2} & \textbf{74.7} & \textbf{71.7} \\  
\rowcolor{gray!20} 
&$_{\pm0.1}$&$_{\pm0.0}$&$_{\pm0.1}$&$_{\pm0.2}$&$_{\pm0.2}$&$_{\pm0.2}$&$_{\pm0.3}$&$_{\pm0.1}$&$_{\pm0.1}$&$_{\pm0.2}$&$_{\pm0.1}$&$_{\pm0.0}$&$_{\pm0.2}$&$_{\pm0.1}$&$_{\pm0.1}$&$_{\pm0.1}$ \\
\midrule
\multicolumn{17}{c}{Level 5} \\
\midrule
Source & 9.4  & 6.7  & 8.3  & 29.1  & 23.4  & 34.0  & 27.0  & 15.8  & 26.3  & 47.4  & 54.7  & 43.9  & 30.5  & 44.5  & 47.6  & 29.9 \\
MEMO~\cite{zhang2022memo} & 21.6  & 17.4  & 20.6  & 37.1  & 29.6  & 40.6  & 34.4  & 25.0  & 34.8  & 55.2  & 65.0  & 54.9  & 37.4  & 55.5  & 57.7  & 39.1 \\
DDA~\cite{gao2022back} & 41.3 & 41.3 & 40.6 & 24.6 & 27.4 & 30.7 & 26.9 & 18.2 & 27.7 & 34.8 & 50.0 & 32.3 & 42.2 & 52.5 & 52.7 & 36.2 \\
TENT~\cite{wang2020tent} & 32.7  & 1.4  & 34.6  & 54.4  & 52.3  & 58.2  & 52.2  & 7.7  & 12.0  & 69.3  & 76.1  & 66.1  & 56.7  & 69.4  & 66.4  & 47.3 \\
% EATA \cite{niu2022efficient} & 35.9  & 34.6  & 36.7  & 45.3  & 47.2  & 49.3  & 47.7  & 56.5  & 55.4  & 62.2  & 72.2  & 21.7  & 56.2  & 64.7  & 63.7  & 49.9  \\
SAR~\cite{niutowards} & 46.5 & 43.1 & 48.9 & \textbf{55.3} & 54.3 & \textbf{58.9} & \textbf{54.8} & 53.6 & 46.2 & 69.7 & \textbf{76.2} & 66.2 & 60.9 & \textbf{69.6} & 66.6 & 58.0\\ 
\rowcolor{gray!20}
\textbf{Ours} & \textbf{47.7} & \textbf{47.1} & \textbf{49.0} & 55.2 & \textbf{54.4} & \textbf{58.9} & 54.7 & \textbf{61.4} & \textbf{54.1} & \textbf{70.1} & 76.0 & \textbf{66.3} & \textbf{61.3} & \textbf{69.6} & \textbf{66.7} & \textbf{59.5} \\  
\rowcolor{gray!20}
&$_{\pm0.4}$&$_{\pm0.4}$&$_{\pm0.4}$&$_{\pm0.1}$&$_{\pm0.1}$&$_{\pm0.3}$&$_{\pm0.1}$&$_{\pm0.1}$&$_{\pm4.8}$&$_{\pm0.1}$&$_{\pm0.1}$&$_{\pm0.1}$&$_{\pm0.2}$&$_{\pm0.1}$&$_{\pm0.2}$&$_{\pm0.4}$ \\
\bottomrule
\end{tabular}
}
\vspace{-10pt}
\end{center}
\end{table*}

\section{Experiments}
\label{sec: Experiments}
In this section, we conduct experiments on benchmark datasets with different test-time adaptation settings, including normal, imbalanced label shifts, and batch size = 1 to evaluate the performance of our proposed VCT method.

\paragraph{Benchmark Datasets and Baselines} 
As in existing papers, we choose the \textbf{ImageNet-C} benchmark dataset \cite{hendrycks2018benchmarking} with $50,000$ instance in $1,000$ classes, with severity level 3 and level 5. 
We also use \textbf{ImageNet-R} \cite{hendrycks2021many}, a dataset containing 30,000 images with various artistic renditions of 200 ImageNet classes.
We evaluate the TTA performance on benchmark datasets for domain adaptation and domain generalization, including \textbf{VisDA-2021} \cite{bashkirova2022visda}, a NeurIPS competition dataset designed to evaluate models' capacity to adapt to novel test distributions and manage distributional shifts and \textbf{Office-Home} which encompasses $15,500$ images with a total of $65$ object categories across four distinct domains. 

We compare our proposed VCT method against the following fully test-time adaptation methods: (1) \textbf{Source:} the baseline model is trained only on the source data without any fine-tuning during the test process.
(2) \textbf{MEMO} \cite{zhang2022memo}: it optimizes the entropy of the averaged prediction over multiple random augmentations of the input sample.
(3) \textbf{DDA} \cite{gao2022back}: it performs input adaptation at test time via a diffusion model.
(4) \textbf{TENT}~\cite{wang2020tent}: it fine-tunes scale and bias parameters of the batch normalization layers using an entropy minimization loss during inference.
(5) \textbf{SAR} \cite{niutowards}: it encourages the model to lie in a flat minimum of the entropy loss surface with a reliable sampling strategy.

% \paragraph{Implementation Details} 
% Following the official implementations of SAR\footnote{SAR: {https://github.com/mr-eggplant/SAR}}, we use the ViT-B/16 backbone for all experiments unless explicitly stated otherwise. We use the pre-trained model weights from the \textit{timm} repository \cite{rw2019timm}. For Office-Home, we fine-tuning the ViT-B/16 by replacing the original classifier head as the source model. For fair performance comparisons, all methods in each experimental setting share the same architecture and the same pre-trained model parameters. We use the SGD optimizer with the Sharpness Aware Minimization \cite{foret2021sharpnessaware}. The batch size is set to 64 for all experiments except the setting \textit{Batch size = 1}. The learning rates of the long-term domain-specific token and the short-term instance-specific token are set to 0.005 and 0.01, respectively. All of our experimental results are the mean values obtained by 3 times running with different random seeds, and the standard deviation is also reported on the last row of tables. All models are trained and tested on a single NVIDIA RTX3090 GPU.

\begin{table*}[!htbp]
\setlength{\abovecaptionskip}{0.1cm}
\setlength{\belowcaptionskip}{-0.2cm}
\begin{center}
\caption{Top-1 Classification Accuracy (\%) for each corruption in \textbf{ImageNet-C} under \textbf{Batch size=1} setting.} % at the highest severity (Level 5).
\label{table: bs1}
\resizebox{\linewidth}{!}
{
\begin{tabular}{l|ccccccccccccccc|c}
\toprule
Method & gaus & shot & impul & defcs & gls & mtn & zm & snw & frst & fg & brt & cnt & els & px & jpg & Avg.\\	
\midrule
\multicolumn{17}{c}{Level 3} \\
\midrule
Source & 51.6 & 46.9 & 50.5 & 48.7 & 37.2 & 54.7 & 41.6 & 35.1 & 33.5 & 67.8 & 69.3 & 74.8 & 65.8 & 66.0 & 63.7 & 53.8 \\
MEMO~\cite{zhang2022memo} & 61.9 & 57.7 & 61.4 & 57.0 & 45.4 & 61.8 & 49.8 & 46.6 & 43.1 & 73.9 & 75.7 & 79.6 & 72.6 & 72.1 & 70.5 & 61.9\\
DDA~\cite{gao2022back} & 59.8 & 58.2 & 59.5 & 43.4 & 43.2 & 50.4 & 40.9 & 34.2 & 34.3 & 55.2 & 64.9 & 64.0 & 64.2 & 63.7 & 62.8 & 53.2\\
TENT~\cite{wang2020tent} & 67.1 & 66.2 & 66.3 & 66.3 & 60.9 & 69.1 & 61.4 & 65.2 & 60.4 & 75.2 & \textbf{78.1} & 78.8 & 74.9 & \textbf{75.8} & 72.4 & 69.2 \\
% EATA \cite{niu2022efficient} & 29.7  & 25.1  & 34.6  & 44.7  & 39.2  & 48.3  & 42.4  & 37.5  & 45.9  & 60.0  & 65.9  & 61.2  & 46.4  & 58.2  & 59.6  & 46.6  \\
SAR~\cite{niutowards} & 68.5  &67.8 & 68.0 & 67.8 & 63.1 & 70.7 & 63.5 & 66.9 & 62.8 &75.8 & 77.7 & 78.4 & 74.7 & 75.7 & 72.7 & 70.3\\ 
\rowcolor{gray!20}
\textbf{Ours} & \textbf{69.6} & \textbf{68.4} & \textbf{68.6} & \textbf{68.5} & \textbf{65.4} & \textbf{71.4} & \textbf{65.1} & \textbf{67.2} & \textbf{67.1} & \textbf{76.0} & 76.4 & \textbf{79.1} & \textbf{75.7} & 75.1 & \textbf{74.4} & \textbf{71.2} \\  
\rowcolor{gray!20}
&$_{\pm0.1}$&$_{\pm0.8}$&$_{\pm0.8}$&$_{\pm0.2}$&$_{\pm1.0}$&$_{\pm0.1}$&$_{\pm0.1}$&$_{\pm1.2}$&$_{\pm0.5}$&$_{\pm0.3}$&$_{\pm0.1}$&$_{\pm0.0}$&$_{\pm0.5}$&$_{\pm0.4}$&$_{\pm0.7}$&$_{\pm0.2}$ \\
\midrule
\multicolumn{17}{c}{Level 5} \\
\midrule
Source & 9.5 & 6.7 & 8.2 & 29.0 & 23.4 & 33.9 & 27.1 & 15.9 & 26.5 & 47.2 & 54.7 & 44.1 & 30.5 & 44.5 & 47.8 & 29.9\\
MEMO~\cite{zhang2022memo} & 21.6  & 17.3  & 20.6  & 37.1  & 29.6  & 40.4  & 34.4  & 24.9  & 34.7  & 55.1  & 64.8  & 54.9  & 37.4  & 55.4  & 57.6  & 39.1\\
DDA~\cite{gao2022back} & 41.3 & \textbf{41.1} & 40.7 & 24.4 & 27.2 & 30.6 & 26.9 & 18.3 & 27.5 & 34.6 & 50.1 & 32.4 & 42.3 & 52.2 & 52.6 & 36.1\\
TENT~\cite{wang2020tent} & 42.2  & 1.0  & 43.3  & 52.4  & 48.2  & 55.5  & 50.5  & 16.5  & 16.9  & 66.4  & \textbf{74.9}  & 64.7  & 51.6  & 67.0  & 64.3  & 47.7 \\
% EATA \cite{niu2022efficient} & 29.7  & 25.1  & 34.6  & 44.7  & 39.2  & 48.3  & 42.4  & 37.5  & 45.9  & 60.0  & 65.9  & 61.2  & 46.4  & 58.2  & 59.6  & 46.6  \\
SAR~\cite{niutowards} & 40.8 & 36.4 & 41.5 & 53.7 & 50.7 & 57.5 & 52.8 & \textbf{59.1} & 50.7 & 68.0 & 74.6 & 65.7 & 57.9 & 68.9 & 65.9 & 56.3\\ 
\rowcolor{gray!20}
\textbf{Ours} & \textbf{46.8} & 40.4 & \textbf{47.7} & \textbf{55.0} & \textbf{55.0} & \textbf{59.5} & \textbf{55.4} & 38.6 & \textbf{62.2} & \textbf{71.0} & 73.7 & \textbf{67.5} & \textbf{63.6} & \textbf{70.4} & \textbf{67.0} & \textbf{58.2} \\  
\rowcolor{gray!20}
&$_{\pm0.1}$&$_{\pm3.5}$&$_{\pm0.3}$&$_{\pm0.1}$&$_{\pm0.1}$&$_{\pm0.1}$&$_{\pm0.2}$&$_{\pm0.8}$&$_{\pm1.7}$&$_{\pm1.0}$&$_{\pm0.2}$&$_{\pm0.0}$&$_{\pm0.1}$&$_{\pm0.1}$&$_{\pm0.8}$&$_{\pm0.2}$ \\
\bottomrule
\end{tabular}
}
% \vspace{-10pt}
\end{center}
\end{table*}

\begin{table*}[!htbp]
\setlength{\abovecaptionskip}{0.1cm}
\setlength{\belowcaptionskip}{-0.2cm}
\begin{center}
\caption{Top-1 Classification Accuracy (\%) for test-time adaptation in \textbf{Office-Home} dataset.}
\label{table: officehome}
\resizebox{\linewidth}{!}
{
\begin{tabular}{l|ccccccccccccccc|r}
\toprule
Methods & A$\rightarrow$C  & A$\rightarrow$P & A$\rightarrow$R & C$\rightarrow$A & C$\rightarrow$P & C$\rightarrow$R & P$\rightarrow$A & P$\rightarrow$C & P$\rightarrow$R & R$\rightarrow$A & R$\rightarrow$C & R$\rightarrow$P & Avg.\\			  
\midrule
Source & 63.4 & 81.9 & 86.3 & 76.2 & 80.6 & 83.8 & 75.0 & 57.9 & 87.2 & 78.7 & 61.0 & 88.0 & 76.7 \\
SAR~\cite{niutowards} & 67.3 & 80.7 & 85.6 & 77.5 & 79.8 & 84.1 & 74.7 & 60.3 & 87.6 & 78.9 & 63.1 & 87.7 & 77.3 \\
\rowcolor{gray!20}
\textbf{Ours} & \textbf{69.1} & \textbf{82.7} & \textbf{87.1} & \textbf{78.4} & \textbf{83.6} & \textbf{85.9} & \textbf{76.5} & \textbf{65.1} & \textbf{87.9} & \textbf{80.2} & \textbf{66.7} & \textbf{88.2} & \textbf{79.3} \\
\rowcolor{gray!20}
&$_{\pm0.0}$&$_{\pm0.0}$&$_{\pm0.0}$&$_{\pm0.0}$&$_{\pm0.0}$&$_{\pm0.0}$&$_{\pm0.0}$&$_{\pm0.0}$&$_{\pm0.0}$&$_{\pm0.0}$&$_{\pm0.0}$&$_{\pm0.0}$&$_{\pm0.0}$ \\
\bottomrule
\end{tabular}
}
% \vspace{-10pt}
\end{center}
\end{table*}

\begin{table*}[!ht]
\setlength{\abovecaptionskip}{0.1cm}
\setlength{\belowcaptionskip}{-0.2cm}
    \centering
    \caption{Top-1 Classification Accuracy (\%)  in \textbf{ImageNet-C} with \textbf{ViT-L/16} under \textbf{Normal} and \textbf{Batch sieze = 1} settings at the highest severity (Level 5).}
    \label{tab: vit-l-normal}
    \resizebox{\linewidth}{!}{
\begin{tabular}{l|ccccccccccccccc|c}
\toprule
Methods & gaus & shot & impul & defcs & gls & mtn & zm & snw & frst & fg & brt & cnt & els & px & jpg & Avg.\\
\midrule
\multicolumn{17}{c}{\textbf{Normal}} \\
    \midrule
        Source & 45.7 & 49.6 & 45.9 & 45.6 & 38.5 & 54.9 & 47.6 & 58.1 & 55.1 & 57.3 & 78.7 & 55.9 & 47.2 & 69.3 & 68.1 & 54.5\\
        TENT & 61.2 & 62.1 & 43.9 & \textbf{59.2} & 49.6 & \textbf{63.3} & \textbf{58.3} & 65.8 & 60.4 & 67.8 & \textbf{80.3} & 37.6 & 60.4 & 74.9 & 72.6 & 61.2 \\
        SAR \cite{niutowards}& 60.7 & 61.2 & 61.0 & 59.1 & 56.8 & 61.6 & 58.2 & 66.6 & 63.2 & \textbf{68.3} & \textbf{80.3} & 63.8 & 60.7 & 74.6 & 69.7 & 64.4 \\
        \rowcolor{gray!20}
        \textbf{Ours} & \textbf{62.0} & \textbf{62.4} & \textbf{62.0} & 57.9 & \textbf{56.9} & 63.1 & \textbf{58.3} & \textbf{67.0} & \textbf{63.6} & 67.1 & \textbf{80.3} & \textbf{68.1} & \textbf{61.5} & \textbf{75.0} & \textbf{72.7} & \textbf{65.2} \\
        \rowcolor{gray!20}  
        &$_{\pm0.1}$&$_{\pm0.1}$&$_{\pm0.1}$&$_{\pm2.0}$&$_{\pm0.9}$&$_{\pm0.8}$&$_{\pm0.2}$&$_{\pm0.1}$&$_{\pm0.3}$&$_{\pm1.3}$&$_{\pm0.2}$&$_{\pm0.1}$&$_{\pm0.1}$&$_{\pm0.2}$&$_{\pm0.0}$&$_{\pm0.2}$ \\
\midrule
\multicolumn{17}{c}{\textbf{Batch size=1}} \\
\midrule
     Source & 45.7 & 49.6 & 45.9 & 45.6 & 38.5 & 54.9 & 47.6 & 58.1 & 55.1 & 57.3 & 78.7 & 55.9 & 47.2 & 69.3 & 68.1 & 54.5\\
     TENT & 62.7 & 63.9 & 62.2 & \textbf{60.7} & 60.3 & \textbf{64.7} & 58.5 & \textbf{68.7} & 41.4 & 68.0 & \textbf{80.8} & 58.6 & 64.2 & 57.6 & \textbf{74.1} & 63.1 \\
    SAR \cite{niutowards}& 62.6 & 62.2 & 62.7 & 53.1 & 52.1 & \textbf{64.7} & 60.0 & 66.9 & 65.3 & 68.4 & 79.5 & 67.6 & 63.7 & \textbf{73.9} & 72.4 & 65.0 \\
    \rowcolor{gray!20}
    \textbf{Ours} & \textbf{64.9} & \textbf{64.2} & \textbf{63.7} & 58.2 & \textbf{61.1} & 57.3 & \textbf{62.3} & 68.0 & \textbf{65.7} & \textbf{68.8} & 79.7 & \textbf{68.1} & \textbf{65.8} & 73.0 & 72.4 & \textbf{66.2} \\
    \rowcolor{gray!20}
    &$_{\pm0.1}$&$_{\pm1.2}$&$_{\pm1.4}$&$_{\pm3.5}$&$_{\pm2.2}$&$_{\pm9.9}$&$_{\pm0.2}$&$_{\pm0.1}$&$_{\pm0.3}$&$_{\pm1.4}$&$_{\pm0.1}$&$_{\pm0.9}$&$_{\pm1.4}$&$_{\pm2.0}$&$_{\pm0.4}$&$_{\pm0.8}$ \\
    \bottomrule
    \end{tabular}   }
\end{table*}

\subsection{Performance Results} 
To evaluate the effectiveness of our proposed approach, we evaluate the performance of our VCT method under three different test conditions on the \textbf{ImageNet-C} dataset with corruption severity level 3 and level 5. Following the SAR method \cite{niutowards}, we report the top-1 accuracy.
As discussed in the implementation details, all methods under comparisons use the same transformer encoder. 
(1) We first evaluate our approach under the \textbf{Normal} i.i.d assumption and compared it with other TTA methods. The results of this experiment are shown in Table \ref{table: normal}.
On average, at both corruption severities, our method outperforms the second-best method. Particularly in level 5, our approach improves the second-best method by 1\%. The ``Ours" row in all tables includes the standard deviation values. These values are generally small, indicating that our method achieves consistent performance across various random seeds.
(2) Then, we evaluate our approach under the \textbf{Imbalanced label shifts} test condition (i.e. the label distribution is long-tailed in a mini-batch) with the same imbalanced sample sequence. The results of this experiment are shown in Table \ref{table: imbalanced}. We can see that our method improves the average classification accuracy of all 15 corruption types by 1.5\% at the highest corruption level. 
(3) We evaluate our approach under the challenging \textbf{Batch size = 1} test condition, which is known to be particularly difficult for TTA methods. As shown in Table \ref{table: bs1}, 
our method improves the average classification accuracy in two corruption severity levels by 0.9\% and 1.9\%,
demonstrating its superior robustness and adaptability with small batch sizes. 
In some categories, our method achieves suboptimal performance. On one side, the DDA method effectively denoises specific corruptions like ``shot noise" using powerful diffusion models. On the other side, the learning rate sensitivity increases when the batch size is reduced to 1. We apply a uniform learning rate to get the best average performance across categories, which may not be ideal for all categories individually.

\begin{table}[!t]
\setlength{\abovecaptionskip}{0.1cm}
\setlength{\belowcaptionskip}{-0.2cm}
    \centering
    \caption{Top-1 Classification Accuracy (\%)  in \textbf{ImageNet-R} and \textbf{VisDA-2021} datasets.}
    \label{tab: imagenet-r}
    \resizebox{0.7\linewidth}{!}{
    \begin{tabular}{l|cc}
    \toprule
        Method &  ImageNet-R & VisDA-2021\\
    \midrule
        Source &  43.1&  44.3\\
        TENT \cite{wang2020tent}& 44.9 & 35.8\\
        SAR \cite{niutowards}& 52.2 & 50.6 \\
        \rowcolor{gray!20}
        \textbf{Ours} & \textbf{53.6} & \textbf{51.1} \\
        \rowcolor{gray!20}
        &$_{\pm0.1}$&$_{\pm0.1}$ \\
    \bottomrule
    \end{tabular}   }
\vspace{-15pt}
\end{table}

We also conducted experiments on the \textbf{ImageNet-R} and \textbf{VisDA-2021} datasets to verify the effectiveness of our method. We use the same pre-trained ViT-B/16 backbone and set the output size to 200 following the procedure in \cite{hendrycks2021many}. From  Table \ref{tab: imagenet-r}, we can see that the overall results are consistent with those on ImageNet-C. The proposed VCT method outperforms the SAR method by 1.4\% and 0.5\% respectively.
For the \textbf{Office-Home} dataset, we also choose the same ViT-B/16 backbone, and the results are presented in Table \ref{table: officehome}. The proposed VCT method outperforms the SAR method by 2.0\%. It demonstrates that the proposed VCT method is effective in different datasets.

We extend our experimentation to encompass a larger ViT-L/16 backbone, operating within the contexts of both Normal and Batch size = 1 settings. For the Normal setting, the batch size is set to 16 due to the limited memory. The results, as illustrated in Table \ref{tab: vit-l-normal}, consistently showcase the superiority of our proposed VCT method over the baseline SAR in both configurations. This robust performance demonstrates the efficacy of our proposed VCT method across diverse transformer backbones.
Overall, our experimental results demonstrate the effectiveness and robustness of our proposed VCT approach in handling complex test conditions and outperforming state-of-the-art TTA methods across multiple evaluation metrics.

\vspace{-0.5cm}
\subsection{Visualization and Analysis}

\begin{figure*}[!ht]
\setlength{\abovecaptionskip}{0.1cm}
\setlength{\belowcaptionskip}{-0.4cm}
    \centering
    \includegraphics[width = 0.9\textwidth]{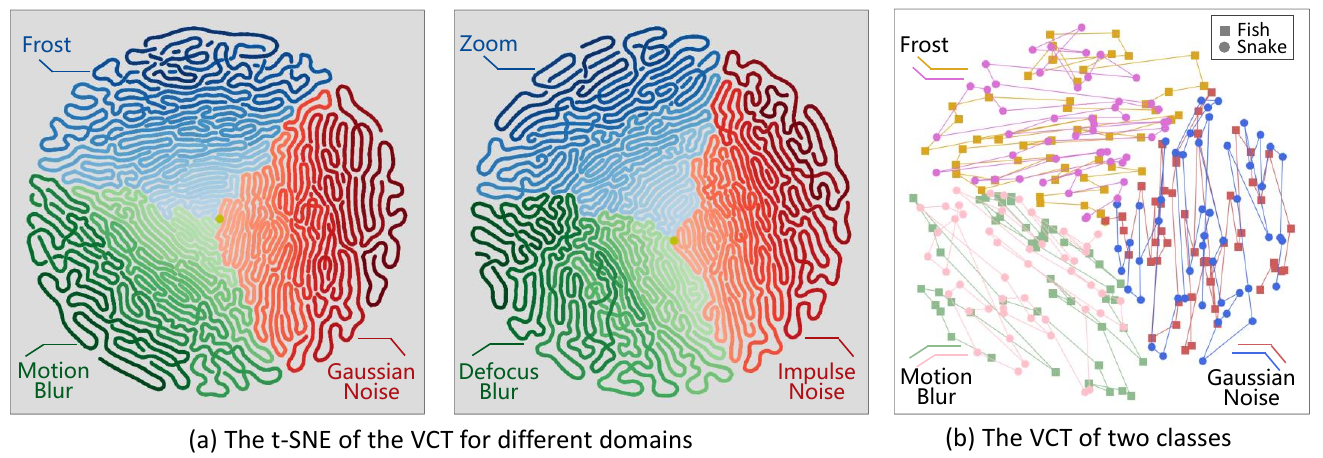}
    \caption{The visualization of the VCT in the adaptation process for different domains of the ImageNet-C dataset. (a): The t-SNE of the VCT for different domains during test-time adaptation, which shows the VCT can learn domain-specific information; (b): The VCT of two classes for different domains during the adaptation process. For the same domain, the VCT cluster within similar zones,  which shows the VCT remains relatively invariant across different categories.}
    \label{fig: visualization}
% \vspace{-0.5cm}
\end{figure*}

Fig. \ref{fig: visualization} (a) shows six example domains in the ImageNet-C dataset of VCT learning during test-time adaptation where the center point represents VCT $\mathbf{C}_0$ obtained from the source model.
Starting from this center point  $\mathbf{C}_0$ provided by the source model, during test-time adaptation, we use $J$ sequential batches of test samples (indexed by $j$) from the target domain (indexed by $K$) to update the VCT $\mathbf{C}_0$, resulting in a sequence 
of $\{\mathbf{C}_0^{j, K}\}$. 
Using the t-SNE method, we project the VCT sequence $\{\mathbf{C}_0^{j, K}| 1\le j\le J\}$ into a 2-D domain and plot the corresponding curve.  Fig. \ref{fig: visualization} (a) shows the VCT curve for 6 different domains. For each domain, the brightness level of the curve represents the batch index $j$ using the sequence of sample batches 
$\{\mathbf{B}_1, \mathbf{B}_2, ..., \mathbf{B}_J\}$
from one of the six target domains. As the adaptation process progresses, the visualization points become increasingly darker.
From these two interesting figures, we can see that, in different domains, the VCT is evolving towards different zones or directions in the vector space. More importantly, the evolution curves of the VCT for different target domains do not overlap with each other. In Fig. \ref{fig: visualization} (b), we choose two classes (Fish and Snake) and only plot the VCTs for samples from these two classes. The square curves represent the VCT vectors for samples from Class Fish and the dot curves represent the VCT vectors for samples from Class Snake. We can see that (1) for samples of the same class but from different domains, their VCTs evolve in totally different directions. (2) For samples from the same domain although from different classes, their VCT curves interleave with each other and evolve in similar zones.
This demonstrates that the VCT is able to capture the domain-specific characteristics of the different target categories for the same domain. These observations motivate us to introduce our proposed method of learning visual conditioning tokens to gradually correct domain shift for fully test-time adaptation.

\begin{figure*}[!ht]
\setlength{\abovecaptionskip}{0.0cm}
\setlength{\belowcaptionskip}{-0.3cm}
    \centering
    \includegraphics[width = 0.97\textwidth]{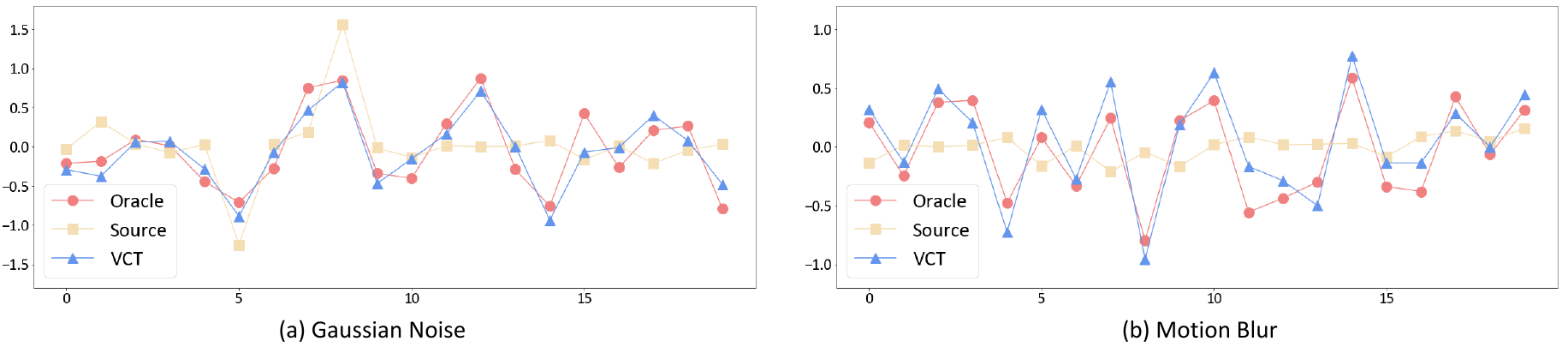}
    \caption{The token comparison of the Source, VCT, and Oracle methods in Gaussian Noise and Motion Blur of ImageNet-C dataset. Our learned VCT is close to the Oracle which is learned by label supervision.}
    \label{fig:numerical}
\end{figure*}

Fig. \ref{fig:numerical} plots the VCT vector learned using the following three methods: (1) the \textbf{Source} method without any adaptation, (2) our proposed \textbf{VCT} method for test-time adaptation without using the sample labels, and (3) the \textbf{Oracle} method where the VCT is learned with ground-truth labels. We can see that the VCT learned using our method is very close to the oracle vector learned with ground-truth labels.
This further demonstrates the effectiveness of our proposed VCT method, enabling robust test-time adaptation and enhancing the model's ability to capture both global domain-specific knowledge and local instance-specific information.

\subsection{Ablation Studies}
Our proposed VCT method has two major components, the domain-specific VCT aiming to capture the domain-specific characteristics and the instance-specific VCT aiming to capture the instance-specific characteristics of the distribution shift.
In this ablation study, we study their contributions to the overall performance. We use the ImageNet-C dataset under the Imbalanced label shifts settings. 
As shown in Table \ref{table: ablation}, when the long-term domain-specific VCT is added to the baseline SAR algorithm, the average accuracy is improved by 0.8\%. When the short-term instance-specific VCT is added, the accuracy is improved by 1.2\%.
When both domain-specific and instance-specific VCTs are added but no reset for the instance-specific VCT, the accuracy is improved by 0.9\% which is lower than the reset version.
When both VCTs are added, our algorithm improves the accuracy by 1.5\% for the experimental setting.
These results demonstrate the importance of incorporating both long-term and short-term tokens to effectively learn both domain-specific and instance-specific information in the target domain.

% \end{minipage}
% \hfill
% \begin{minipage}{0.57\textwidth}
\begin{table}[!t]
    \centering
    \caption{Ablation study under \textbf{Imbalanced label shifts} in ImageNet-C dataset at the highest severity (Level 5). }
\label{table: ablation}
\resizebox{0.6\linewidth}{!}
{
\begin{tabular}{l|cc}
\toprule
Methods  & Avg. \\	
\midrule
% Source & 29.9 & 29.9\\
Baseline  Method & 58.0 \\
\quad + Domain-Specific VCT  & 58.8\\
\quad + Instance-Specific VCT  & 59.1\\
Our Method without Reset & 58.9\\
\rowcolor{gray!20}
\textbf{Our Method}  & \textbf{59.5}\\
\bottomrule
\end{tabular}
}
% % \end{minipage}
% \vspace{-0.5cm}
\end{table}

\begin{figure}
\setlength{\abovecaptionskip}{0.1cm}
\setlength{\belowcaptionskip}{-0.5cm}
    \centering
    % \vspace{-10pt}
    \includegraphics[width=0.9\linewidth]{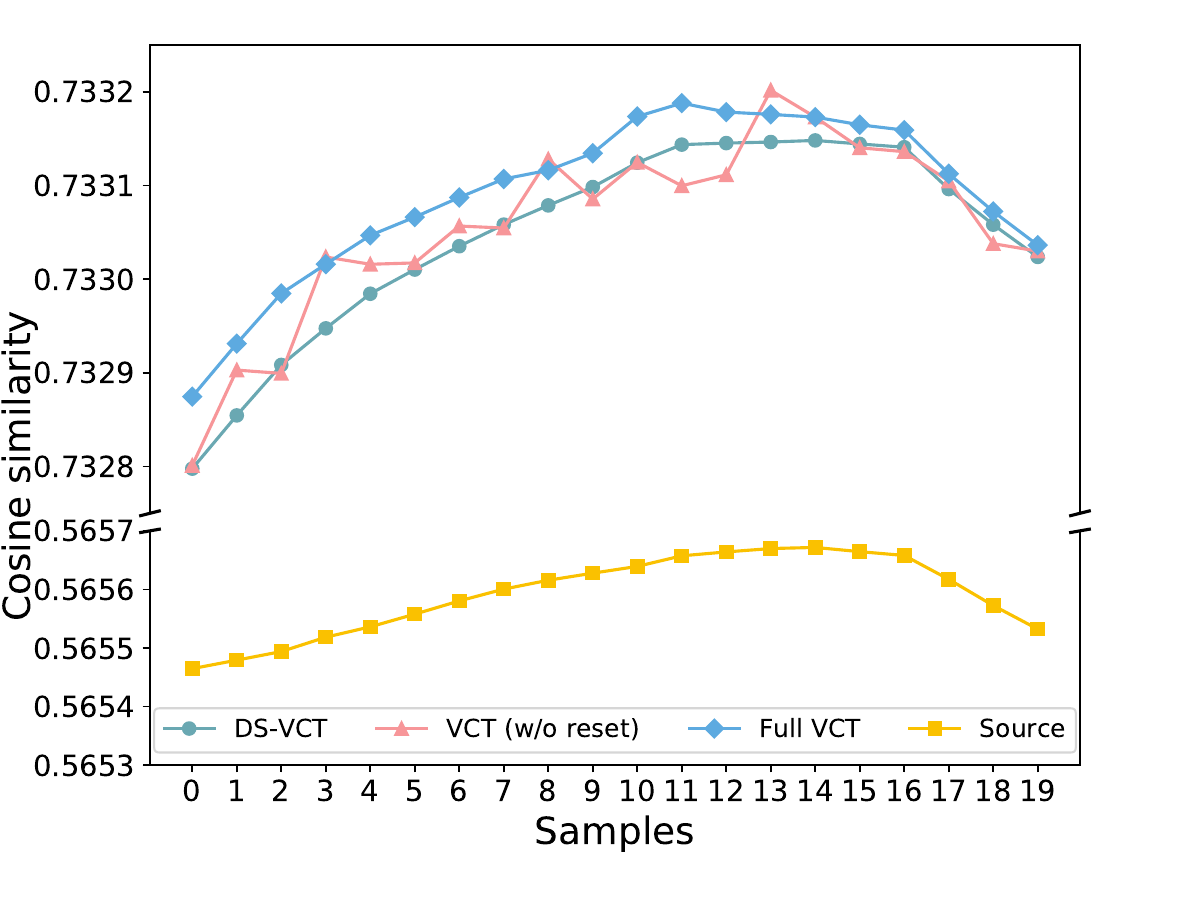}
    \caption{The cosine similarity between the label-supervised Oracle token and our distinct token configurations.}\label{fig: similarity}
\end{figure}

We use the following example to demonstrate the need to reset the instance-specific VCT after each batch. 
Let the Oracle VCT be the VCT learned with labeled samples during testing. 
In Fig. \ref{fig: similarity}, we plot the cosine similarity values between the Oracle VCT and the following three VCTs:  (a) the domain-specific VCT labeled with DS-VCT, (b) the VCT obtained by our method with reset of the IS-VCT (labeled with Full VCT),  (c) the VCT without reset (labeled with VCT w/o reset), and (d) the original VCT from the source model.
We can see that the VCT with reset (Full VCT) is much closer to the Oracle VCT than the source model VCT. Without the reset of IS-VCT, the cosine similarity becomes unstable.  This demonstrates that the reset operation for IS-VCT is effective.

\begin{figure}[!ht]
\setlength{\abovecaptionskip}{0.0cm}
\setlength{\belowcaptionskip}{-0.4cm}
    \centering
    \includegraphics[width=\linewidth]{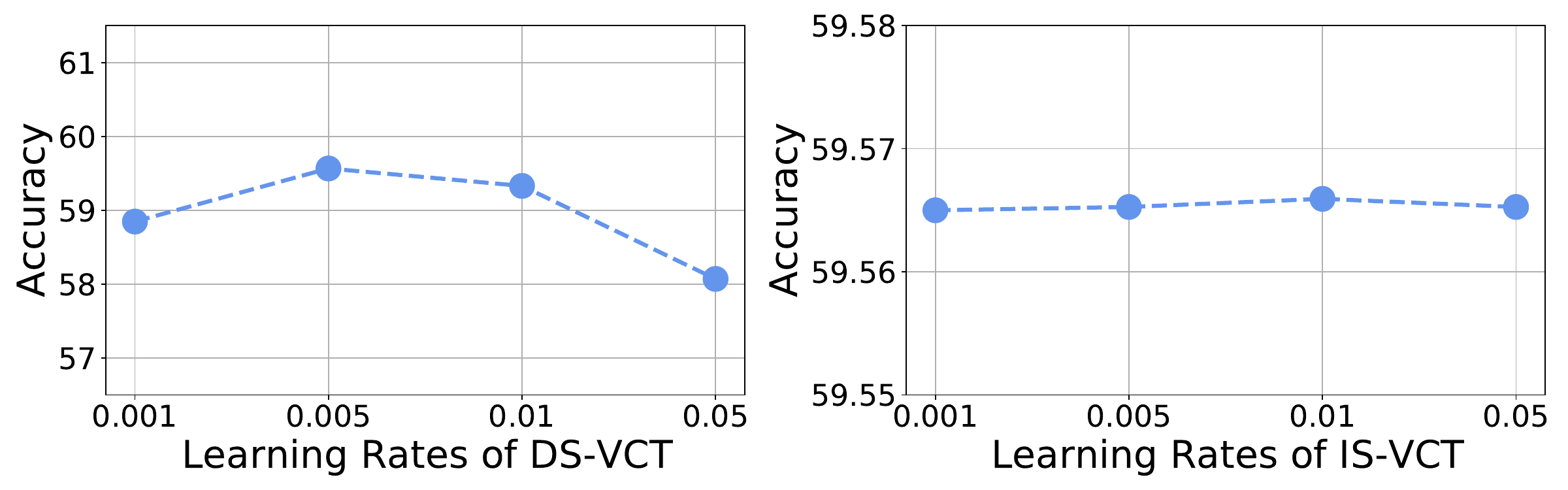}
    \caption{Sensitivity analyses for the learning rates of DS-VCT and IS-VCT.}
    \label{fig:sensitivity}
\end{figure}

We also conduct parameter sensitivity analyses on the learning rates for the DS-VCT and IS-VCT with the Imbalanced label shifts setting in ImageNet-C. When evaluating one hyper-parameter, the others are fixed to their default values. As shown in Fig. \ref{fig:sensitivity}, we can see that the performance is best when the learning rate of DS-VCT is set to $0.005$. Furthermore, it achieves stable performance within the range of the learning rate of IS-VCT between $0.001$ and $0.05$. Therefore, we set the learning rate to $0.005$ and $0.01$ for the DS-VCT and IS-VCT respectively for all experiments.

\section{Conclusion and Further Discussions}
\label{sec: conclution}
Fully test-time adaptation is a challenging problem in computer vision, particularly in the presence of complex corruptions and shifts in the test data distribution. 
We have made an interesting finding that the class token at the first layer of the transformer encoder can be used as a target for optimization during test-time adaptation. 
We have developed a bi-level approach to effectively learn the visual conditioning token. 
Our experimental results demonstrate that our proposed bi-level visual conditioning token learning method is able to achieve significantly improved test-time adaptation performance.

While the DS-VCT is effective at capturing domain-specific prior knowledge and has shown promising results in many single-domain settings, its effectiveness can be limited in the presence of mixed-domain test data. In this test condition, our VCT method can only improve the accuracy of the baseline SAR method by 0.1\%. Specifically, when the test data consists of a mixture of domains, the DS-VCT may not be able to effectively model the complex interactions between the domains and may result in performance degradations. This is because the DS-VCT is designed to capture long-term domain-specific information of the target samples. The presence of mixed-domain data can create conflicts between these long-term domain-specific priors, leading to degraded generalization performance. 
In our future work, we aim to address this limitation by encoding the mix-domain targets to a unified latent space based on diffusion models.

\section{Acknowledgment}
This work was supported by Project 2021JC02X103 and Center for Computational Science and Engineering at Southern University of Science and Technology.

\bibliographystyle{IEEEtran}
\bibliography{IEEEabrv,ref}

\begin{thebibliography}{10}
\providecommand{\url}[1]{#1}
\csname url@samestyle\endcsname
\providecommand{\newblock}{\relax}
\providecommand{\bibinfo}[2]{#2}
\providecommand{\BIBentrySTDinterwordspacing}{\spaceskip=0pt\relax}
\providecommand{\BIBentryALTinterwordstretchfactor}{4}
\providecommand{\BIBentryALTinterwordspacing}{\spaceskip=\fontdimen2\font plus
\BIBentryALTinterwordstretchfactor\fontdimen3\font minus \fontdimen4\font\relax}
\providecommand{\BIBforeignlanguage}[2]{{%
\expandafter\ifx\csname l@#1\endcsname\relax
\typeout{** WARNING: IEEEtran.bst: No hyphenation pattern has been}%
\typeout{** loaded for the language `#1'. Using the pattern for}%
\typeout{** the default language instead.}%
\else
\language=\csname l@#1\endcsname
\fi
#2}}
\providecommand{\BIBdecl}{\relax}
\BIBdecl

\bibitem{quinonero2008dataset}
J.~Quinonero-Candela, M.~Sugiyama, A.~Schwaighofer, and N.~D. Lawrence, \emph{Dataset shift in machine learning}.\hskip 1em plus 0.5em minus 0.4em\relax MIT Press, 2008.

\bibitem{mirza2021robustness}
M.~J. Mirza, C.~Buerkle, J.~Jarquin, M.~Opitz, F.~Oboril, K.-U. Scholl, and H.~Bischof, ``Robustness of object detectors in degrading weather conditions,'' in \emph{2021 IEEE International Intelligent Transportation Systems Conference (ITSC)}.\hskip 1em plus 0.5em minus 0.4em\relax IEEE, 2021, pp. 2719--2724.

\bibitem{wang2022cross}
R.~Wang, Z.~Wu, Z.~Weng, J.~Chen, G.-J. Qi, and Y.-G. Jiang, ``Cross-domain contrastive learning for unsupervised domain adaptation,'' \emph{IEEE Transactions on Multimedia}, 2022.

\bibitem{active_zhou}
S.~Zhou, L.~Wang, S.~Zhang, Z.~Wang, and W.~Zhu, ``Active gradual domain adaptation: Dataset and approach,'' \emph{IEEE Transactions on Multimedia}, vol.~24, pp. 1210--1220, 2022.

\bibitem{guided_zhang}
C.~Zhang, Z.~Li, J.~Liu, P.~Peng, Q.~Ye, S.~Lu, T.~Huang, and Y.~Tian, ``Self-guided adaptation: Progressive representation alignment for domain adaptive object detection,'' \emph{IEEE Transactions on Multimedia}, vol.~24, pp. 2246--2258, 2022.

\bibitem{meng2022dual}
M.~Meng, M.~Lan, J.~Yu, J.~Wu, and L.~Liu, ``Dual-level adaptive and discriminative knowledge transfer for cross-domain recognition,'' \emph{IEEE Transactions on Multimedia}, 2022.

\bibitem{deng2023cross}
J.~Deng, X.~Zhang, W.~Li, L.~Duan, and D.~Xu, ``Cross-domain detection transformer based on spatial-aware and semantic-aware token alignment,'' \emph{IEEE Transactions on Multimedia}, 2023.

\bibitem{chen2024learning}
S.~Chen, Y.~Tang, Z.~Kan, and Z.~He, ``Learning inference-time drift sensor-actuator for domain generalization,'' in \emph{ICASSP 2024-2024 IEEE International Conference on Acoustics, Speech and Signal Processing (ICASSP)}.\hskip 1em plus 0.5em minus 0.4em\relax IEEE, 2024, pp. 5090--5094.

\bibitem{zhang2024hiker}
C.~Zhang, S.~Stepputtis, J.~Campbell, K.~Sycara, and Y.~Xie, ``Hiker-sgg: Hierarchical knowledge enhanced robust scene graph generation,'' in \emph{Proceedings of the IEEE/CVF Conference on Computer Vision and Pattern Recognition}, 2024, pp. 28\,233--28\,243.

\bibitem{liang2020we}
J.~Liang, D.~Hu, and J.~Feng, ``Do we really need to access the source data? source hypothesis transfer for unsupervised domain adaptation,'' in \emph{International Conference on Machine Learning}.\hskip 1em plus 0.5em minus 0.4em\relax PMLR, 2020, pp. 6028--6039.

\bibitem{Wang_2022_CVPR}
F.~Wang, Z.~Han, Y.~Gong, and Y.~Yin, ``Exploring domain-invariant parameters for source free domain adaptation,'' in \emph{Proceedings of the IEEE/CVF Conference on Computer Vision and Pattern Recognition}, June 2022, pp. 7151--7160.

\bibitem{li2020model}
R.~Li, Q.~Jiao, W.~Cao, H.-S. Wong, and S.~Wu, ``Model adaptation: Unsupervised domain adaptation without source data,'' in \emph{Proceedings of the IEEE/CVF Conference on Computer Vision and Pattern Recognition}, 2020, pp. 9641--9650.

\bibitem{limttn}
H.~Lim, B.~Kim, J.~Choo, and S.~Choi, ``Ttn: A domain-shift aware batch normalization in test-time adaptation,'' in \emph{The Eleventh International Conference on Learning Representations}, 2023.

\bibitem{tang2023cross}
Y.~Tang, Q.~Guo, and Z.~He, ``Cross-inferential networks for source-free unsupervised domain adaptation,'' in \emph{2023 IEEE International Conference on Image Processing (ICIP)}.\hskip 1em plus 0.5em minus 0.4em\relax IEEE, 2023, pp. 96--100.

\bibitem{tian2022vdmda}
J.~Tian, J.~Zhang, W.~Li, and D.~Xu, ``Vdm-da: Virtual domain modeling for source data-free domain adaptation,'' \emph{IEEE Transactions on Circuits and Systems for Video Technology}, vol.~32, no.~6, pp. 3749--3760, 2022.

\bibitem{Sunttt}
Y.~Sun, X.~Wang, Z.~Liu, J.~Miller, A.~A. Efros, and M.~Hardt, ``Test-time training with self-supervision for generalization under distribution shifts,'' in \emph{{International Conference on Machine Learning}}, vol. 119.\hskip 1em plus 0.5em minus 0.4em\relax {PMLR}, 2020, pp. 9229--9248.

\bibitem{wang2020tent}
D.~Wang, E.~Shelhamer, S.~Liu, B.~Olshausen, and T.~Darrell, ``Tent: Fully test-time adaptation by entropy minimization,'' in \emph{International Conference on Learning Representations}, 2020.

\bibitem{mirza2022norm}
M.~J. Mirza, J.~Micorek, H.~Possegger, and H.~Bischof, ``The norm must go on: Dynamic unsupervised domain adaptation by normalization,'' in \emph{Proceedings of the IEEE/CVF Conference on Computer Vision and Pattern Recognition}, 2022, pp. 14\,765--14\,775.

\bibitem{niu2022efficient}
S.~Niu, J.~Wu, Y.~Zhang, Y.~Chen, S.~Zheng, P.~Zhao, and M.~Tan, ``Efficient test-time model adaptation without forgetting,'' in \emph{International Conference on Machine Learning}.\hskip 1em plus 0.5em minus 0.4em\relax PMLR, 2022, pp. 16\,888--16\,905.

\bibitem{wang2022continual}
Q.~Wang, O.~Fink, L.~Van~Gool, and D.~Dai, ``Continual test-time domain adaptation,'' in \emph{Proceedings of the IEEE/CVF Conference on Computer Vision and Pattern Recognition}, 2022, pp. 7201--7211.

\bibitem{tang2023neuro}
Y.~Tang, C.~Zhang, H.~Xu, S.~Chen, J.~Cheng, L.~Leng, Q.~Guo, and Z.~He, ``Neuro-modulated hebbian learning for fully test-time adaptation,'' in \emph{Proceedings of the IEEE/CVF Conference on Computer Vision and Pattern Recognition}, 2023, pp. 3728--3738.

\bibitem{yuan2023robust}
L.~Yuan, B.~Xie, and S.~Li, ``Robust test-time adaptation in dynamic scenarios,'' in \emph{Proceedings of the IEEE/CVF Conference on Computer Vision and Pattern Recognition}, 2023, pp. 15\,922--15\,932.

\bibitem{wen2023test}
Z.~Wen, S.~Niu, G.~Li, Q.~Wu, M.~Tan, and Q.~Wu, ``Test-time model adaptation for visual question answering with debiased self-supervisions,'' \emph{IEEE Transactions on Multimedia}, 2023.

\bibitem{kan2023self}
Z.~Kan, S.~Chen, C.~Zhang, Y.~Tang, and Z.~He, ``Self-correctable and adaptable inference for generalizable human pose estimation,'' in \emph{Proceedings of the IEEE/CVF Conference on Computer Vision and Pattern Recognition}, 2023, pp. 5537--5546.

\bibitem{liu2021ttt++}
Y.~Liu, P.~Kothari, B.~van Delft, B.~Bellot-Gurlet, T.~Mordan, and A.~Alahi, ``Ttt++: When does self-supervised test-time training fail or thrive?'' in \emph{Advances in Neural Information Processing Systems}, vol.~34, 2021, pp. 21\,808--21\,820.

\bibitem{gandelsman2022test}
Y.~Gandelsman, Y.~Sun, X.~Chen, and A.~Efros, ``Test-time training with masked autoencoders,'' \emph{Advances in Neural Information Processing Systems}, vol.~35, pp. 29\,374--29\,385, 2022.

\bibitem{niutowards}
S.~Niu, J.~Wu, Y.~Zhang, Z.~Wen, Y.~Chen, P.~Zhao, and M.~Tan, ``Towards stable test-time adaptation in dynamic wild world,'' in \emph{International Conference on Learning Representations}, 2023.

\bibitem{zhang2022memo}
M.~Zhang \emph{et~al.}, ``Memo: Test time robustness via adaptation and augmentation,'' \emph{Advances in Neural Information Processing Systems}, vol.~35, pp. 38\,629--38\,642, 2022.

\bibitem{jing2022variational}
M.~Jing, X.~Zhen, J.~Li, and C.~Snoek, ``Variational model perturbation for source-free domain adaptation,'' \emph{Advances in Neural Information Processing Systems}, vol.~35, pp. 17\,173--17\,187, 2022.

\bibitem{he2016deep}
K.~He, X.~Zhang, S.~Ren, and J.~Sun, ``Deep residual learning for image recognition,'' in \emph{Proceedings of the IEEE conference on computer vision and pattern recognition}, 2016, pp. 770--778.

\bibitem{zagoruyko2016wide}
S.~Zagoruyko and N.~Komodakis, ``Wide residual networks,'' in \emph{British Machine Vision Conference 2016}.\hskip 1em plus 0.5em minus 0.4em\relax British Machine Vision Association, 2016.

\bibitem{vaswani2017attention}
A.~Vaswani, N.~Shazeer, N.~Parmar, J.~Uszkoreit, L.~Jones, A.~N. Gomez, {\L}.~Kaiser, and I.~Polosukhin, ``Attention is all you need,'' \emph{Advances in neural information processing systems}, vol.~30, 2017.

\bibitem{dosovitskiyimage}
A.~Dosovitskiy, L.~Beyer, A.~Kolesnikov, D.~Weissenborn, X.~Zhai, T.~Unterthiner, M.~Dehghani, M.~Minderer, G.~Heigold, S.~Gelly \emph{et~al.}, ``An image is worth 16x16 words: Transformers for image recognition at scale,'' in \emph{International Conference on Learning Representations}, 2021.

\bibitem{tang2022gloss}
S.~Tang, R.~Hong, D.~Guo, and M.~Wang, ``Gloss semantic-enhanced network with online back-translation for sign language production,'' in \emph{Proceedings of the 30th ACM International Conference on Multimedia}, 2022, pp. 5630--5638.

\bibitem{he2022revisiting}
J.~He, R.~Hong, X.~Liu, M.~Xu, and Q.~Sun, ``Revisiting local descriptor for improved few-shot classification,'' \emph{ACM Transactions on Multimedia Computing, Communications, and Applications (TOMM)}, vol.~18, no.~2s, pp. 1--23, 2022.

\bibitem{jia2022visual}
M.~Jia, L.~Tang, B.-C. Chen, C.~Cardie, S.~Belongie, B.~Hariharan, and S.-N. Lim, ``Visual prompt tuning,'' in \emph{European Conference on Computer Vision}.\hskip 1em plus 0.5em minus 0.4em\relax Springer, 2022, pp. 709--727.

\bibitem{gan2023decorate}
Y.~Gan, Y.~Bai, Y.~Lou, X.~Ma, R.~Zhang, N.~Shi, and L.~Luo, ``Decorate the newcomers: Visual domain prompt for continual test time adaptation,'' in \emph{Proceedings of the AAAI Conference on Artificial Intelligence}, vol.~37, no.~6, 2023, pp. 7595--7603.

\bibitem{DBLP:conf/emnlp/LesterAC21}
B.~Lester, R.~Al{-}Rfou, and N.~Constant, ``The power of scale for parameter-efficient prompt tuning,'' in \emph{{Proceedings of the 2021 Conference on Empirical Methods in Natural Language Processing}}, 2021, pp. 3045--3059.

\bibitem{liu2021ptuningv2}
X.~Liu, K.~Ji, Y.~Fu, W.~L. Tam, Z.~Du, Z.~Yang, and J.~Tang, ``P-tuning v2: Prompt tuning can be comparable to fine-tuning universally across scales and tasks,'' \emph{arXiv preprint arXiv:2110.07602}, 2021.

\bibitem{li2021prefix}
X.~L. Li and P.~Liang, ``Prefix-tuning: Optimizing continuous prompts for generation,'' \emph{arXiv preprint arXiv:2101.00190}, 2021.

\bibitem{hendrycks2018benchmarking}
D.~Hendrycks and T.~Dietterich, ``Benchmarking neural network robustness to common corruptions and perturbations,'' in \emph{International Conference on Learning Representations}, 2018.

\bibitem{mcclelland1995there}
J.~L. McClelland \emph{et~al.}, ``Why there are complementary learning systems in the hippocampus and neocortex: insights from the successes and failures of connectionist models of learning and memory.'' \emph{Psychological review}, vol. 102, no.~3, p. 419, 1995.

\bibitem{kumaran2016learning}
D.~Kumaran, D.~Hassabis, and J.~L. McClelland, ``What learning systems do intelligent agents need? complementary learning systems theory updated,'' \emph{Trends in cognitive sciences}, vol.~20, no.~7, pp. 512--534, 2016.

\bibitem{DBLP:conf/icml/SunWLMEH20}
Y.~Sun, X.~Wang, Z.~Liu, J.~Miller, A.~A. Efros, and M.~Hardt, ``Test-time training with self-supervision for generalization under distribution shifts,'' in \emph{{ICML}}, vol. 119.\hskip 1em plus 0.5em minus 0.4em\relax {PMLR}, 2020, pp. 9229--9248.

\bibitem{choi2022improving}
S.~Choi, S.~Yang, S.~Choi, and S.~Yun, ``Improving test-time adaptation via shift-agnostic weight regularization and nearest source prototypes,'' \emph{arXiv preprint arXiv:2207.11707}, 2022.

\bibitem{marsden2022gradual}
R.~A. Marsden, M.~D{\"o}bler, and B.~Yang, ``Gradual test-time adaptation by self-training and style transfer,'' \emph{arXiv preprint arXiv:2208.07736}, 2022.

\bibitem{boudiaf2022parameter}
M.~Boudiaf, R.~Mueller, I.~Ben~Ayed, and L.~Bertinetto, ``Parameter-free online test-time adaptation,'' in \emph{Proceedings of the IEEE/CVF Conference on Computer Vision and Pattern Recognition}, 2022, pp. 8344--8353.

\bibitem{liu2023promptsurvey}
P.~Liu, W.~Yuan, J.~Fu, Z.~Jiang, H.~Hayashi, and G.~Neubig, ``Pre-train, prompt, and predict: A systematic survey of prompting methods in natural language processing,'' \emph{ACM Computing Surveys}, vol.~55, no.~9, pp. 1--35, 2023.

\bibitem{zhou2022coop}
K.~Zhou, J.~Yang, C.~C. Loy, and Z.~Liu, ``Learning to prompt for vision-language models,'' \emph{International Journal of Computer Vision}, vol. 130, no.~9, pp. 2337--2348, 2022.

\bibitem{zhou2022conditional}
K.~Zhou, J.~Yang, C.~Loy, Chen, and Z.~Liu, ``Conditional prompt learning for vision-language models,'' in \emph{Proceedings of the IEEE/CVF Conference on Computer Vision and Pattern Recognition}, 2022, pp. 16\,816--16\,825.

\bibitem{ma2023prompt}
C.~Ma, Y.~Liu, J.~Deng, L.~Xie, W.~Dong, and C.~Xu, ``Understanding and mitigating overfitting in prompt tuning for vision-language models,'' \emph{IEEE Transactions on Circuits and Systems for Video Technology}, vol.~33, no.~9, pp. 4616--4629, 2023.

\bibitem{shu2022test}
M.~Shu, W.~Nie, D.-A. Huang, Z.~Yu, T.~Goldstein, A.~Anandkumar, and C.~Xiao, ``Test-time prompt tuning for zero-shot generalization in vision-language models,'' in \emph{Advances in Neural Information Processing Systems}, 2022.

\bibitem{ge2022DAPL}
C.~Ge, R.~Huang, M.~Xie, Z.~Lai, S.~Song, S.~Li, and G.~Huang, ``Domain adaptation via prompt learning,'' \emph{arXiv preprint arXiv:2202.06687}, 2022.

\bibitem{zhang2024concept}
Y.~Zhang, C.~Zhang, K.~Yu, Y.~Tang, and Z.~He, ``Concept-guided prompt learning for generalization in vision-language models,'' \emph{arXiv preprint arXiv:2401.07457}, 2024.

\bibitem{zhang2024cross}
Y.~Zhang, C.~Zhang, Y.~Tang, and Z.~He, ``Cross-modal concept learning and inference for vision-language models,'' \emph{Neurocomputing}, vol. 583, p. 127530, 2024.

\bibitem{zhang2023bdc}
Y.~Zhang, C.~Zhang, Z.~Liao, Y.~Tang, and Z.~He, ``Bdc-adapter: Brownian distance covariance for better vision-language reasoning,'' \emph{arXiv preprint arXiv:2309.01256}, 2023.

\bibitem{radford2021clip}
A.~Radford, J.~W. Kim, C.~Hallacy, A.~Ramesh, G.~Goh, S.~Agarwal, G.~Sastry, A.~Askell, P.~Mishkin, J.~Clark \emph{et~al.}, ``Learning transferable visual models from natural language supervision,'' in \emph{International Conference on Machine Learning}.\hskip 1em plus 0.5em minus 0.4em\relax PMLR, 2021, pp. 8748--8763.

\bibitem{oh2023blackvip}
C.~Oh, H.~Hwang, H.-y. Lee, Y.~Lim, G.~Jung, J.~Jung, H.~Choi, and K.~Song, ``Blackvip: Black-box visual prompting for robust transfer learning,'' in \emph{Proceedings of the IEEE/CVF Conference on Computer Vision and Pattern Recognition}, 2023, pp. 24\,224--24\,235.

\bibitem{sohn2022visualgan}
K.~Sohn, Y.~Hao, J.~Lezama, L.~Polania, H.~Chang, H.~Zhang, I.~Essa, and L.~Jiang, ``Visual prompt tuning for generative transfer learning,'' \emph{arXiv preprint arXiv:2210.00990}, 2022.

\bibitem{gao2022visual}
Y.~Gao, X.~Shi, Y.~Zhu, H.~Wang, Z.~Tang, X.~Zhou, M.~Li, and D.~N. Metaxas, ``Visual prompt tuning for test-time domain adaptation,'' \emph{arXiv preprint arXiv:2210.04831}, 2022.

\bibitem{parisi2019continual}
G.~I. Parisi, R.~Kemker, J.~L. Part, C.~Kanan, and S.~Wermter, ``Continual lifelong learning with neural networks: A review,'' \emph{Neural networks}, vol. 113, pp. 54--71, 2019.

\bibitem{o2004division}
R.~C. O’Reilly, ``The division of labor between the neocortex and hippocampus,'' in \emph{Connectionist models in cognitive psychology}.\hskip 1em plus 0.5em minus 0.4em\relax Psychology Press, 2004, pp. 153--162.

\bibitem{gais2007sleep}
S.~Gais, G.~Albouy, M.~Boly, T.~T. Dang-Vu, A.~Darsaud, M.~Desseilles, G.~Rauchs, M.~Schabus, V.~Sterpenich, G.~Vandewalle \emph{et~al.}, ``Sleep transforms the cerebral trace of declarative memories,'' \emph{Proceedings of the National Academy of Sciences}, vol. 104, no.~47, pp. 18\,778--18\,783, 2007.

\bibitem{kitamura2017engrams}
T.~Kitamura, S.~K. Ogawa, D.~S. Roy, T.~Okuyama, M.~D. Morrissey, L.~M. Smith, R.~L. Redondo, and S.~Tonegawa, ``Engrams and circuits crucial for systems consolidation of a memory,'' \emph{Science}, vol. 356, no. 6333, pp. 73--78, 2017.

\bibitem{aranilearning}
E.~Arani, F.~Sarfraz, and B.~Zonooz, ``Learning fast, learning slow: A general continual learning method based on complementary learning system,'' in \emph{International Conference on Learning Representations}, 2022.

\bibitem{yu2023continual}
D.~Yu, M.~Zhang, M.~Li, F.~Zha, J.~Zhang, L.~Sun, and K.~Huang, ``Contrastive correlation preserving replay for online continual learning,'' \emph{IEEE Transactions on Circuits and Systems for Video Technology}, pp. 1--1, 2023.

\bibitem{neves2008synaptic}
G.~Neves, S.~F. Cooke, and T.~V. Bliss, ``Synaptic plasticity, memory and the hippocampus: a neural network approach to causality,'' \emph{Nature Reviews Neuroscience}, vol.~9, no.~1, pp. 65--75, 2008.

\bibitem{zhang2021token}
H.~Zhang, Y.~Hao, and C.-W. Ngo, ``Token shift transformer for video classification,'' in \emph{Proceedings of the 29th ACM International Conference on Multimedia}, 2021, pp. 917--925.

\bibitem{xu2022multi}
L.~Xu, W.~Ouyang, M.~Bennamoun, F.~Boussaid, and D.~Xu, ``Multi-class token transformer for weakly supervised semantic segmentation,'' in \emph{Proceedings of the IEEE/CVF Conference on Computer Vision and Pattern Recognition}, 2022, pp. 4310--4319.

\bibitem{gao2022back}
J.~Gao, J.~Zhang, X.~Liu, T.~Darrell, E.~Shelhamer, and D.~Wang, ``Back to the source: Diffusion-driven test-time adaptation,'' in \emph{2023 IEEE/CVF Conference on Computer Vision and Pattern Recognition}, 2023.

\bibitem{hendrycks2021many}
D.~Hendrycks, S.~Basart, N.~Mu, S.~Kadavath, F.~Wang, E.~Dorundo, R.~Desai, T.~Zhu, S.~Parajuli, M.~Guo, D.~Song, J.~Steinhardt, and J.~Gilmer, ``The many faces of robustness: A critical analysis of out-of-distribution generalization,'' \emph{Proceedings of the IEEE/CVF international conference on computer vision}, 2021.

\bibitem{bashkirova2022visda}
D.~Bashkirova, D.~Hendrycks, D.~Kim, H.~Liao, S.~Mishra, C.~Rajagopalan, K.~Saenko, K.~Saito, B.~U. Tayyab, P.~Teterwak \emph{et~al.}, ``Visda-2021 competition: Universal domain adaptation to improve performance on out-of-distribution data,'' in \emph{NeurIPS 2021 Competitions and Demonstrations Track}.\hskip 1em plus 0.5em minus 0.4em\relax PMLR, 2022, pp. 66--79.

\end{thebibliography}

\begin{IEEEbiography}[{\includegraphics[width=1in,height=1.25in,clip,keepaspectratio]{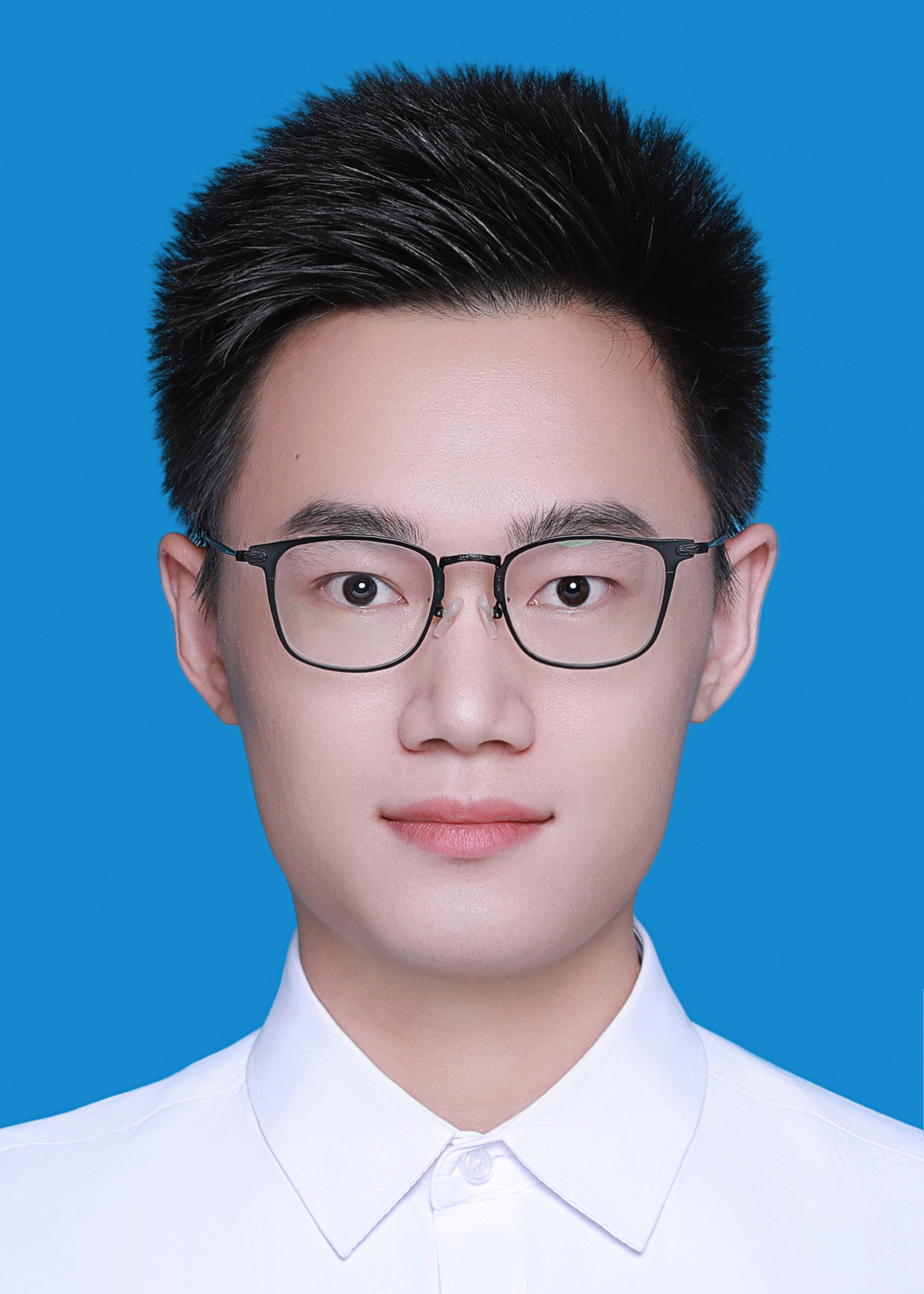}}]{Yushun Tang}
is currently pursuing a Ph.D. in Intelligent Manufacturing and Robotics at the Southern University of Science and Technology (SUSTech). He holds a Bachelor’s degree in Optoelectronic Information Science and Engineering from Harbin Engineering University and a Master’s degree in Electronic Science and Technology from SUSTech. His current research focuses on Computer Vision, Transfer Learning, and Domain Adaptation.
\end{IEEEbiography}

% \vspace{-1cm}
\begin{IEEEbiography}[{\includegraphics[width=1in,height=1.25in,clip,keepaspectratio]{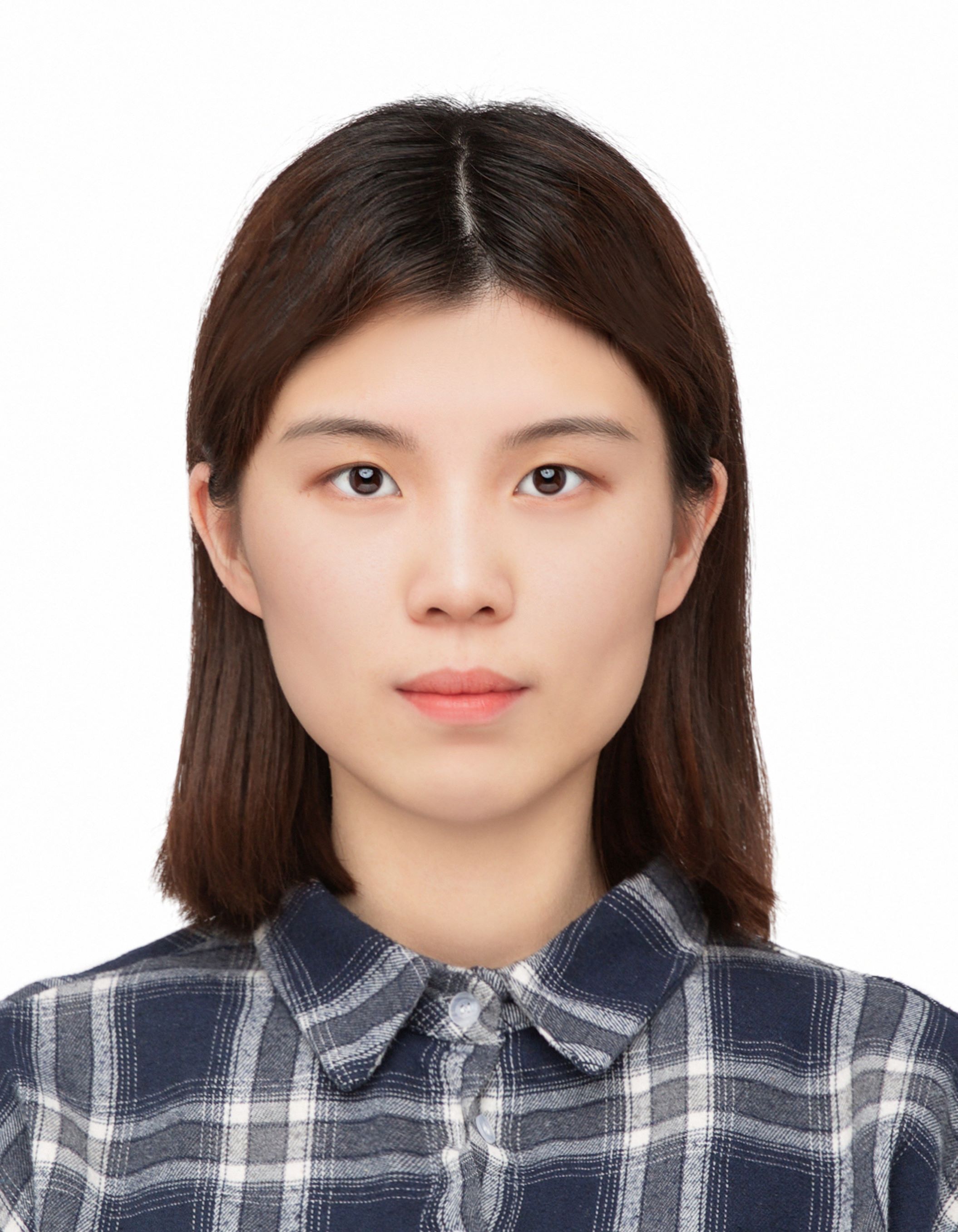}}]{Shuoshuo Chen}
received the B.Eng. degree from Southern University of Science and Technology, Shenzhen, China, in 2021. She is currently pursuing the M.Eng. degree with the Department of Electrical and Electronic Engineering, Southern University of Science and Technology, Shenzhen, China.
Her research interests include the generalization and adaptation capability of deep learning models. 
\end{IEEEbiography}

% \vspace{-1cm}
\begin{IEEEbiography}
[{\includegraphics[width=1in,height=1.25in,clip,keepaspectratio]{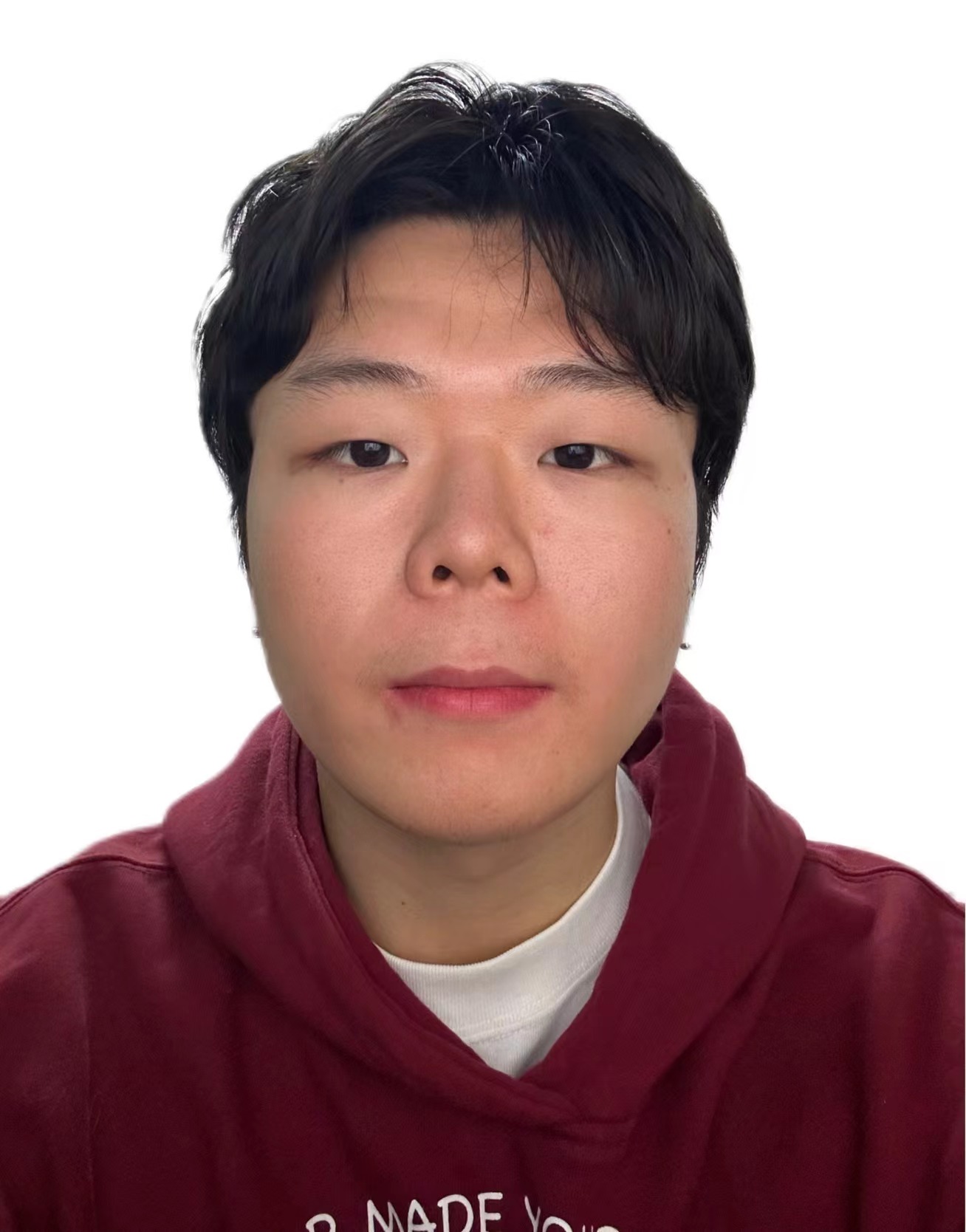}}]{Zhehan Kan}
received the B.S. degree from the School of Computer Science and Information Engineering, Hefei University of Technology, Hefei, China, in 2021, and the M.S. degree in the Department of Electrical and Electronic Engineering, Southern University of Science and Technology, Shenzhen, China, in 2023. He is currently working toward the Ph.D. degree in Electronic and Information Engineering with Tsinghua University, Beijing, China. His research interests include computer vision and multimodal large language model.
\end{IEEEbiography}

% \vspace{-1cm}
\begin{IEEEbiography}
[{\includegraphics[width=1in,height=1.25in,clip,keepaspectratio]{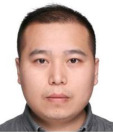}}]{Yi Zhang}
is currently pursuing a Ph.D. in Artificial Intelligence at the Southern University of Science and Technology (SUSTech) and Harbin Institute of Technology, with an expected graduation date in July 2025. Prior to this, he earned a Bachelor’s degree in Software Engineering from Northeastern University (China) and a Master’s degree in Information Systems from The University of Texas. His current research focuses on vision-language models, Few-shot Learning, and Visual Reasoning.
\end{IEEEbiography}

\begin{IEEEbiography}
[{\includegraphics[width=1in,height=1.25in,clip,keepaspectratio]{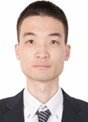}}]{Qinghai Guo} is currently a Principal Engineer at the Advanced Computing and Storage Lab of Huawei Technologies Co., Ltd and the Team Leader of the Brain-inspired Computing Team. He received his Doctoral degree in Mathematical Sciences from the Institute of Mathematical Stochastics, University of Goettingen, Germany in 2017. His research focuses on brain-inspired computing and machine learning, including spiking neural networks, biological plausible and efficient learning theories, neuromorphic computing architecture, etc. 
\end{IEEEbiography}

\begin{IEEEbiography}
[{\includegraphics[width=1in,height=1.25in,clip,keepaspectratio]{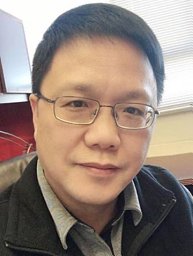}}]{Zhihai He} received the B.S. degree in mathematics from Beijing Normal University, Beijing, China, in 1994, the M.S. degree in mathematics from the Institute of Computational Mathematics, Chinese Academy of Sciences, Beijing, China, in 1997, and the Ph.D. degree in electrical engineering from the University of California, at Santa Barbara, CA, USA, in 2001. In 2001, he joined Sarnoff Corporation, Princeton, NJ, USA, as a member of technical staff. In 2003, he joined the Department of Electrical and Computer Engineering, University of Missouri, Columbia, MO, USA, where he was a tenured full professor. He is currently a chair professor with the Department of Electrical and Electronic Engineering, Southern University of Science and Technology, Shenzhen, P. R. China. His current research interests include image/video processing and compression, wireless sensor network, computer vision, and cyber-physical systems.

He is a member of the Visual Signal Processing and Communication Technical Committee of the IEEE Circuits and Systems Society. He serves as a technical program committee member or a session chair of a number of international conferences. He was a recipient of the 2002 {\sc IEEE Transactions on Circuits and Systems for Video Technology} Best Paper Award and the SPIE VCIP Young Investigator Award in 2004. He was the co-chair of the 2007 International Symposium on Multimedia Over Wireless in Hawaii. He has served as an Associate Editor for the {\sc IEEE Transactions on Circuits and Systems for Video Technology} (TCSVT), the {\sc IEEE Transactions on Multimedia} (TMM), and the Journal of Visual Communication and Image Representation. He was also the Guest Editor for the IEEE TCSVT Special Issue on Video Surveillance.
\end{IEEEbiography}

\end{document}